\documentclass{article}
 \usepackage[preprint]{neurips_2026}

\usepackage[utf8]{inputenc} 
\usepackage[T1]{fontenc}    
\usepackage{hyperref}       
\usepackage{url}            
\usepackage{booktabs}       
\usepackage{amsfonts}       
\usepackage{nicefrac}       
\usepackage{microtype}      
\usepackage{amsmath}
\usepackage{amssymb}
\usepackage{mathtools}
\usepackage{amsthm}
\usepackage[table,xcdraw]{xcolor}
\usepackage{multirow}
\usepackage{wrapfig}

\title{MMGS: \textbf{10$\times$} Compressed 3DGS through Optimal Transport Aggregation based on Multi-view Ranking}

\author{Beizhen Zhao\textsuperscript{1}
\and Sicheng Yu\textsuperscript{1}
\and Ziran Yin\textsuperscript{1} 
\and Dongxu Shen\textsuperscript{1} 
\and Hao Wang\textsuperscript{1\dag}
\and \textsuperscript{1}The Hong Kong University of Science and Technology (Guangzhou)
\and
\quad
{\small \texttt{bzhao610@connect.hkust-gz.edu.cn} \quad} 
{\small \texttt{haowang@hkust-gz.edu.cn}}
\and \textsuperscript{\dag}Corresponding Author
}

\begin{document}

\maketitle

\begin{abstract}
While 3D Gaussian Splatting (3DGS) has revolutionized 3D reconstruction, it suffers from significant overhead due to massive redundant primitives. 
Existing compression methods typically rely on local sampling or fixed pruning thresholds, which often struggle to balance redundancy reduction with high-fidelity rendering. 
To address this, we propose a novel framework that formulates Gaussian optimization as a global geometric distribution matching problem. 
Specifically, our approach integrates three components: 
(1) we introduce a multi-view 3D Gaussian contribution ranking mechanism that filters primitives using geometric consistency instead of local heuristics; 
(2) we propose a global Optimal Transport (OT)-based aggregation algorithm that merges redundant primitives while preserving the underlying geometry; 
and (3) we design an OT-based densification operator that maintains the Gaussian's distributional properties for stable optimization. 
Our approach achieves state-of-the-art rendering quality with only \textbf{10$\%$} primitives and \textbf{10$\times$} accelerated training speeds compared to vanilla 3DGS.

\end{abstract}

\section{Introduction}

3D Gaussian Splatting (3DGS) has revolutionized explicit scene reconstruction with real-time rendering~\cite{fei20243d,wu2024recent}. 
However, this framework relies heavily on 2D view-space gradients to guide the densification process of primitives~\cite{bao20253d,ren2024octree,lu2024scaffold,bagdasarian20253dgs,liu2024atomgs,ali2024trimming,chen2024hac}. 
The ambiguity of 2D view-space supervision leads to massive redundant Gaussians and geometric degeneration. 
To minimize the rendering loss, the optimizer often spawns massive numbers of minute Gaussians, acting as patches to overfit specific viewpoints rather than forming a coherent structure. 
Moreover, gradient-threshold heuristics inevitably result in primitive redundancy.
This dependence on 2D view-space heuristics for 3D structural optimization creates a fundamental challenge, limiting the framework's ability to achieve compact and geometrically accurate representations.

To address this issue, existing works broadly fall into two categories of 3DGS compression methods. 
The first focuses on importance-based sampling, like Taming-3DGS~\cite{mallick2024taming} and FastGS~\cite{ren2025fastgs}, which assess primitive significance using pixel-wise scores and multi-view consistency. 
However, these approaches rely on local sampling heuristics and lack a global compression strategy, often leaving substantial redundancy. 
The second category adopts hard pruning strategies, such as depth downsampling in Mini-splatting~\cite{fang2024mini} and the pruning scheme in Speedy-splat~\cite{hanson2025speedy}. 
Although effective in reducing primitive counts, these methods prune only existing Gaussians and depend on fixed thresholds, limiting adaptability across scenes and frequently degrading rendering quality.

In this work, we propose a novel Gaussian compression framework that integrates multi-view Gaussian contribution ranking and Optimal Transport (OT) theory~\cite{villani2008optimal,thorpe2018introduction,montesuma2024recent,moon1996expectation}. 
Our motivation is to decouple the generation of primitives from 2D projection error, ensuring that every Gaussian exists to represent valid 3D geometry rather than merely overfitting a specific viewpoint.
We treat the optimization process as a geometric distribution matching problem.
By aligning the Gaussian distribution to geometrically valid regions via contribution ranking and global OT aggregation, we enforce sparsity while maintaining structural integrity.

To this end, we first introduce a multi-view ranking mechanism.
Specifically, we construct a texture-aware metric map to filter out transient noise. 
We then project the 2D map back into 3D space via an inverse accumulation voting mechanism and execute importance sampling for densification and pruning.
This ensures that new Gaussians are initialized at geometrically valid locations, preventing floaters and artifacts caused by view-dependent overfitting.

To further address Gaussian redundancy, we model Gaussian reduction as a discrete mass transport problem rooted in OT theory. 
Technically, our algorithm minimizes the Wasserstein distance between the original dense set and a compacted set. 
Through global OT aggregation, we merge spatially clustered low-contribution primitives into a compacted set.
This approach significantly reduces the primitive count while maintaining the geometric continuity of the underlying distribution.

Finally, we propose an OT-based densification strategy. 
Existing methods often duplicate opacity or scale arbitrarily, leading to optimization instability. 
We propose a new splitting operator for stable optimization process, ensuring that when a Gaussian is split, the sub-Gaussians maintain the distributional properties of the parent.

By integrating these modules, our framework achieves superior rendering quality to vanilla 3DGS with only \textbf{10$\%$} primitives and \textbf{10$\times$} accelerated training speeds suitable for scalable applications.
To summarize our contributions:

\begin{itemize}
    \item We propose a multi-view 3D Gaussian ranking mechanism that resolves the geometric ambiguity of 2D gradients by validating candidate Gaussians at consistent 3D locations.

    \item We treat Gaussian redundancy reduction as an optimal transport problem by globally aggregating primitives into representative Gaussians based on ranking score.

    \item We introduce an OT-based densification strategy that preserves the parent distribution during splitting for stable optimization.
\end{itemize}

\section{Related Work}

Existing work has attempted to refine density control to mitigate geometric redundancy.
Taming 3DGS~\cite{mallick2024taming} introduces a budgeting mechanism that ranks Gaussians based on opacity, scale, and accumulation counts, curbing excessive growth. 
Similarly, DashGaussian~\cite{chen2025dashgaussian} employs a resolution-guided scheduler to progressively reconstruct scene geometry. 
To incorporate geometric cues, GaussianPro~\cite{cheng2024gaussianpro} utilizes depth and normal maps to guide propagation.
Kheradmand et al.~\cite{kheradmand20243d} framed 3DGS optimization as a Markov Chain Monte Carlo (MCMC) process, using stochastic updates to migrate Gaussians. 
More recently, FastGS~\cite{ren2025fastgs} has integrated multi-view consistency checks to filter unstable primitives and achieves superb efficiency.
Despite these advancements, most existing densification strategies remain tethered to 2D projection heuristics causing primitives to overfit specific viewpoints.

Parallel to densification, pruning redundant Gaussians also helps accelerate rendering and reduce storage. 
Methods like LightGaussian~\cite{fan2024lightgaussian} and RadSplat~\cite{niemeyer2025radsplat} calculate importance scores to prune insignificant primitives. 
PUP 3DGS~\cite{hanson2025pup} approximates the Hessian trace to guide sensitive pruning. 
Compact3DGS~\cite{lee2024compact} and LP-3DGS~\cite{zhang2024lp} employ learnable masks for redundant Gaussians.
StopThePop~\cite{radl2024stopthepop} and FlashGS~\cite{feng2025flashgs} optimize tile intersection and memory access patterns to speed up inference.
3DGS-LM~\cite{hollein20253dgs} uses Levenberg-Marquardt algorithm for faster convergence, and 3DGS$^2$~\cite{lan20253dgs2} utilizes second-order updates. 
However, pruning often degrades details because it lacks a mechanism to redistribute the information of removed Gaussians into the remaining set. 

While GHAP~\cite{wang2025gaussian} claims that it conceptualizes compaction as an OT problem, its distance metric design is in contradiction to the OT theory.
Based on their implementation, the cost matrix is computed using a direct Euclidean distance and the Frobenius norm of the simple difference between covariance matrices: $C_{ij} = ||\mu_1 - \mu_2||_2^2 + ||\Sigma_1 - \Sigma_2||_F^2$. 
Computing distances via direct linear subtraction ignores non-Euclidean structure. 
To address these challenges, Our proposed framework utilizes multi-view ranking to anchor valid geometry and Optimal Transport theory to minimize distributional redundancy, ensuring a compact yet accurate scene representation.

\begin{figure*}[ht!]
  \vspace{-20pt}
  \begin{center}
    \centerline{\includegraphics[width=0.99\columnwidth]{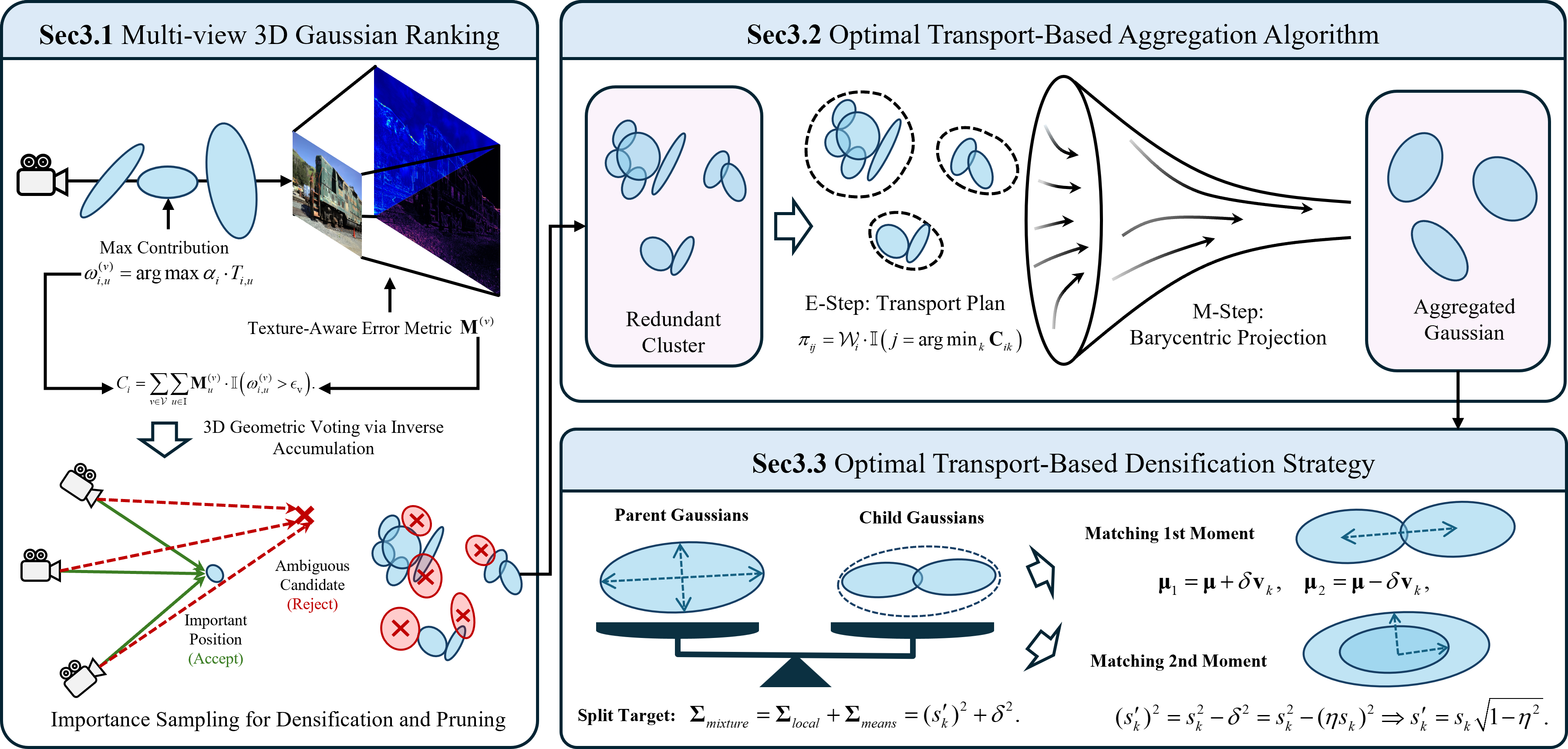}}
    \caption{\textbf{The pipeline of MMGS.} First, we propose a multi-view 3D Gaussian ranking mechanism that resolves the geometric ambiguity of 2D gradients by validating candidate Gaussians at consistent 3D locations.
    Then, we design a global optimal transport algorithm to aggregate primitives into representative Gaussians based on ranking scores.
    Finally, we introduce an OT-based densification strategy that preserves the parent distribution during splitting for stable optimization.}
    \label{frame}
  \end{center}
  \vspace{-20pt}
\end{figure*}

\section{Methodology}

We aim to reconstruct a high-fidelity 3D radiance field with a minimal number of Gaussian primitives $\mathcal{G} = \{g_1, \dots, g_N\}$. 
Our framework treats the optimization process as a geometric distribution matching problem rather than overfitting a specific viewpoint.
Therefore, we propose a pipeline based on multi-view contribution ranking and Optimal Transport.

\subsection{Multi-View 3D Gaussian Ranking}
\label{sec:importance_ranking}

The standard 3DGS densification is based on the gradient with respect to 2D projection loss. 
However, 2D gradient may lead to ambiguous candidate to overfit view-dependent errors.
To address this, we propose a multi-view 3D ranking mechanism. 
This method identifies important Gaussians by detecting structural anomalies across multiple observations, ensuring that densification occurs only in regions with verified geometric inconsistency.

\subsubsection{Texture-Aware Structural map}

We first establish a robust error metric beyond simple photometric differences. 
Pure $L_1$ loss is often insensitive to structural details where Gaussians struggle to represent effectively.
We define a texture-aware structural map $\mathbf{M}$ that captures both gradient intensity and local curvature.

Let $\mathbf{I} \in \mathbb{R}^{H \times W \times 3}$ be an RGB image. 
We first transform it into luminance space $\ell(\mathbf{I})$.
Then, to capture structural edges, we employ a composite operator.
The gradient magnitude map $\mathbf{G}$ is defined as $\mathbf{G}(\ell) = \sqrt{(\mathbf{K}_x * \ell)^2 + (\mathbf{K}_y * \ell)^2 + \epsilon},$
where $\mathbf{K}_x, \mathbf{K}_y$ are the horizontal and vertical Sobel kernels, and $*$ denotes the convolution operator. The curvature map $\mathbf{H}$ is defined via the discrete Laplacian $\mathbf{H}(\ell) = |\mathbf{K}_{Lap} * \ell|.$
The composite texture map $\mathbf{T}(\mathbf{I})$ combines these first-order and second-order derivatives to highlight regions of high geometric complexity:
\begin{equation}
    \mathbf{T}(\mathbf{I}) = \mathcal{N}\left( \mathbf{G}(\ell(\mathbf{I})) + \lambda \mathbf{H}(\ell(\mathbf{I})) \right),
\end{equation}
where $\lambda$ balances edge and corner sensitivity 
, and $\mathcal{N}(\cdot)$ denotes min-max normalization to $[0,1]$.

We then construct a dual-criterion error mask $\mathbf{M}^{(v)}$ for a given view $v$. 
An error pixel is considered valid if and only if it exhibits both significant photometric deviation and structural deviation. 
Let $\hat{\mathbf{I}}^{(v)}$ be the rendered image and $\mathbf{I}^{(v)}$ be the ground truth. The binary mask is defined as:
\begin{equation}
    \mathbf{M}^{(v)} = \mathbb{I}\left( |(\mathbf{T}(\hat{\mathbf{I}}^{(v)}) - \mathbf{T}(\mathbf{I}^{(v)})| > \tau_1 \right) \land \mathbb{I}\left( |\hat{\mathbf{I}}^{(v)} - \mathbf{I}^{(v)}| > \tau_2 \right).
\end{equation}
where $\tau$ is a predefined threshold and $\mathbb{I}(\cdot)$ is the indicator function. 
This intersection encourages to focus on structural misalignment rather than transient noise.

\subsubsection{3D Geometric Voting via Inverse Accumulation}

We use a multi-view voting module to locate specific 3D primitives for geometric consistency. 
Instead of only relying on instantaneous gradient, we accumulate a geometric deficiency score over a subset of viewpoints $\mathcal{V}$.

For a pixel $u$ in view $v$, let $\omega_{i,u}^{(v)} = \alpha_i \cdot T_{i,u}$ represent the contribution weight of Gaussian $g_i$, where $\alpha_i$ denotes opacity and $T_{i,u}$ denotes transmittance on pixel $u$. 
We define the geometric deficiency score $C_i$ for Gaussian $g_i$ by accumulating the error flags across all views:
\begin{equation}
    C_i = \sum_{v \in \mathcal{V}} \sum_{u \in \text{I}} \mathbf{M}^{(v)}_u \cdot \mathbb{I}\left( \omega_{i,u}^{(v)} > \epsilon_{\text{v}} \right).
\end{equation}
A high $C_i$ indicates that $g_i$ is consistently failing to reconstruct geometry across diverse viewing angles, suggesting a further optimization.

\subsubsection{Importance Sampling for Densification and Pruning}

Finally, we derive two distinct metrics to guide the lifecycle of the Gaussians:

\paragraph{Densification Score ($S_d$).} This score identifies regions where the current Gaussian density is insufficient to represent the geometry. 
It is derived from multi-view averaged consistency count $S_{d}(g_i) = \left\lfloor \frac{C_i}{|\mathcal{V}|} \right\rfloor.$
Gaussians with $S_d(g_i) \geq 1$ are marked as geometrically deficient and prioritized for densification.

\paragraph{Pruning Score ($S_{p}$).} A weighted score is used to guide the removal of redundant primitives:
\begin{equation}
    S_p(g_i) = \mathcal{N}( \sum_{v \in \mathcal{V}} \mathcal{L}^{(v)} \cdot \sum_{u} \mathbf{M}^{(v)}_u \cdot \omega_{i,u}^{(v)}),
\end{equation}
where $\mathcal{L} =
(1 - \lambda_{\text{ssim}})\,\mathcal{L}_{\text{L1}}
+ \lambda_{\text{ssim}}\,(1 - \mathcal{L}_\text{SSIM})$.
A lower pruning score indicates that the primitive contributes little to reducing the global photometric error and is thus more likely to be redundant.
This module decouples geometric consistency from photometric error to specifically target hard geometric regions for densification while pruning soft redundant regions.

\subsection{OT-Based Aggregation Algorithm}
\label{sec:ot_aggregation}

While the ranking method in Sec.~\ref{sec:importance_ranking} removes some floaters, it does not resolve the redundancy of overlapping Gaussians that represent the same local surface. 
To further compress redundancy of Gaussian collection, we propose a global aggregation mechanism. 
We formulate Gaussian reduction not as a pruning task, but as a discrete Optimal Transport (OT) problem: finding a sparse set of primitives that minimizes the Wasserstein distance to the original dense distribution.

\subsubsection{Importance-Weighted Spatial Partitioning}

Solving a global OT problem over millions of Gaussians is computationally intractable.
To make it feasible, we employ an adaptive importance-weighted partitioning strategy instead of standard spatial partitioning based on the fact that the information density of a radiance field is non-uniform. 
To render a pixel $u$, 3DGS utilizes rasterization algorithm $ u = \sum_{i \in N} c_i\alpha_i \prod_{j=1}^{i-1} (1 - \alpha_j) = \sum_{i \in N} c_i\alpha_i T_i$ to blend contributed Gaussian collection $N$.

To simulate the rasterization rendering process, we introduce a visual contribution score $\mathcal{W}_i$ for each Gaussian $g_i$, combining its opacity $\alpha_i$ and geometric deficiency score to evaluate the ``mass'' for our transport problem $ \mathcal{W}_i = \alpha_i \cdot C_i.$

We then construct an importance-balanced KD-Tree. 
Unlike standard KD-Tree based on the median index, we split space such that the cumulative visual contribution is balanced:
\begin{equation}
    \sum_{g_i \in \mathcal{P}_{\text{left}}} \mathcal{W}_i \approx \sum_{g_j \in \mathcal{P}_{\text{right}}} \mathcal{W}_j \approx \frac{1}{2} \sum_{g_k \in \mathcal{P}_{\text{parent}}} \mathcal{W}_k.
\end{equation}
This ensures that the reduced blocks $\mathcal{B}_1, \dots, \mathcal{B}_M$ are not spatially equal, but informationally equal, allowing us to allocate a uniform budget to each block locally.

\subsubsection{Gaussian Wasserstein Distance and Cost Matrix}

Within each block $\mathcal{B}$, we aim to map original Gaussians $\mathcal{S}_{src} = \{(\mu_i, \Sigma_i, \alpha_i)\}_{i=1}^N$ to a smaller set $\mathcal{S}_{tgt} = \{(\mu'_j, \Sigma'_j, \alpha'_j)\}_{j=1}^K$, where $K \ll N$.
We measure the distance between two Gaussian distributions through Bures-Wasserstein metric~\cite{ruschendorf1985wasserstein, bhatia2019bures}.
The squared Wasserstein distance between $\mathcal{S}(\mu_i, \Sigma_i)$ and $\mathcal{S}(\mu_j, \Sigma_j)$ is:
\begin{equation}
    d_{W}^2(\mathcal{S}_i, \mathcal{S}_j) = \text{Tr}\left( \Sigma_i + \Sigma_j - 2(\Sigma_i^{1/2} \Sigma_j \Sigma_i^{1/2})^{1/2} \right) + \|\mu_i - \mu_j\|_2^2.
\end{equation}
However, computing the matrix square root of the product $\Sigma_i^{1/2} \Sigma_j \Sigma_i^{1/2}$ for every pair is prohibitively expensive. 
To accelerate this, we employ the Gelbrich Distance~\cite{gelbrich1990formula}, a tight lower bound that serves as an efficient proxy in our optimization target (detailed in Appendix.~\ref{sec:gelbrich_validation}:
\begin{equation}
    d_{G}^2(\mathcal{S}_i, \mathcal{S}_j) = \|\mu_i - \mu_j\|_2^2 + \|\Sigma_i^{1/2} - \Sigma_j^{1/2}\|_F^2,
\end{equation}
where $\|\cdot\|_F$ is the Frobenius norm.
The cost matrix $\mathbf{C} \in \mathbb{R}^{N \times K}$ is efficiently computed:
\begin{equation}
    \mathbf{C}_{ij} =  d_{G}^2(\mathcal{S}_{src}, \mathcal{S}_{tgt}).
\end{equation}

\subsubsection{Barycentric Projection and Aggregation}

We solve the aggregation via an iterative Expectation-Maximization (EM) algorithm rooted in OT.

\paragraph{E-Step (Transport Plan):} We determine the optimal transport plan $\pi \in \mathbb{R}^{N \times K}$ that reallocates the probability mass of the original Gaussians to the target set. 
We employ a hard clustering strategy based on the minimal transport cost. 
The transport plan is calculated as:
\begin{equation}
    \pi_{ij} = \mathcal{W}_i \cdot \mathbb{I}\left(j = \arg\min_{k} \mathbf{C}_{ik}\right),
\end{equation}
where $\mathcal{W}_i$ denotes the importance weight of the $i$-th source primitive, and $\mathbb{I}(\cdot)$ is the indicator function.

\paragraph{M-Step (Moment Matching):} We update the parameters of the target Gaussians to match the full statistical moments of the assigned source cluster. 
This is achieved via strict moment matching, ensuring the aggregated Gaussian preserves the total variance of the cluster:
\begin{equation}
    \mu'_j = \frac{\sum_i \pi_{ij} \mu_i}{\sum_i \pi_{ij}}, 
    \Sigma'_j = \frac{\sum_i \pi_{ij} \left( \Sigma_i + (\mu_i - \mu'_j)(\mu_i - \mu'_j)^T \right)}{\sum_i \pi_{ij}}.
\end{equation}
To efficiently compute $(\mu_i - \mu'_j)(\mu_i - \mu'_j)^T$ without incurring prohibitive memory costs from explicit difference matrices, we implement the update using the variance identity $\text{Var}(X) = \mathbb{E}[X^2] - (\mathbb{E}[X])^2$.
This block-wise OT aggregation allows us to reduce primitive counts by orders of magnitude while minimizing the divergence of the represented 3D probability density function.

\subsection{OT-based Densification Strategies with Contractive Regularization}
\label{sec:densification}

Standard 3DGS densification strategies are heuristic and introduce sudden discontinuities, causing optimization instability and requiring long warm-up periods. 
We propose an OT-based densification strategy for stable optimization.
When a Gaussian $g \sim \mathcal{N}(\boldsymbol{\mu}, \boldsymbol{\Sigma})$ needs to be split, 
we aim to split $g$ into two sub-Gaussians $g_1, g_2$ such that their combined distribution preserves the first and second moments of the original $g$ at the moment of creation.

Let $g_1, g_2$ have means $\boldsymbol{\mu}_1, \boldsymbol{\mu}_2$ and identical covariances $\boldsymbol{\Sigma}'$. 
To preserve the 1st moment, we require $\frac{1}{2}(\boldsymbol{\mu}_1 + \boldsymbol{\mu}_2) = \boldsymbol{\mu}$. 
We define the separation along the principal axis of the original covariance. 
Let $\boldsymbol{\Sigma} = \mathbf{R} \mathbf{S}^2 \mathbf{R}^T$, where $\mathbf{S} = \text{diag}(s_1, s_2, s_3)$. 
We identify the principal axis index $k = \arg\max_i s_i$ and the corresponding eigenvector $\mathbf{v}_k$ (the $k$-th column of $\mathbf{R}$).
The new means are defined as symmetric offsets:
\begin{equation}
    \boldsymbol{\mu}_1 = \boldsymbol{\mu} + \delta \mathbf{v}_k, \quad \boldsymbol{\mu}_2 = \boldsymbol{\mu} - \delta \mathbf{v}_k,
\end{equation}
where $\delta = \eta \cdot s_k$ is the displacement magnitude modulated by a splitting factor $\eta$ (more details in Appendix.~\ref{supp:densification_proof}).

To preserve the 2nd moment, the mixture variance must equal the original variance. 
Along the principal axis, the combined second moment is:
\begin{equation}
    \boldsymbol{\Sigma} _{mixture} = \boldsymbol{\Sigma} _{local} + \boldsymbol{\Sigma} _{means} \implies  (s'_k)^2 + \delta^2.
\end{equation}
Setting this equal to $s_k^2$, we derive the update rule for the scale along the principal axis:
\begin{equation}
    (s'_k)^2 = s_k^2 - (\eta s_k)^2 \implies s'_k = s_k \sqrt{1 - \eta^2}.
\end{equation}

\begin{table*}[t]
\centering
\caption{\textbf{Efficiency and Render Quality Trade-off across Baselines.} We evaluate efficiency by decoupling it into \textbf{Compression Ratio} (the primitive count of vanilla 3DGS divided by the method's primitive count) and \textbf{$\Delta$PSNR} (absolute PSNR shift relative to vanilla 3DGS). A superior method achieves high sparsity while maintaining $\Delta\text{PSNR} \ge 0$. 
}
\vspace{-5pt}
\label{tab:ratio_comparison}
\center
\setlength{\tabcolsep}{5pt} 
\resizebox{1.0\linewidth}{!}{
\begin{tabular}{l|cc|cc|cc}
\toprule
\multirow{2}{*}{Method} & \multicolumn{2}{c|}{Deep Blending} & \multicolumn{2}{c|}{Tanks \& Temples} & \multicolumn{2}{c}{Waymo} \\
& Comp. Ratio $\uparrow$ & $\Delta$PSNR $\uparrow$ & Comp. Ratio $\uparrow$ & $\Delta$PSNR $\uparrow$ & Comp. Ratio $\uparrow$ & $\Delta$PSNR $\uparrow$ \\
\midrule
DashGaussian   & 1.27$\times$ & -0.13 & 1.30$\times$ & +0.18 & 1.43$\times$ & -1.27   \\
Speedy-splat   & 5.13$\times$ & -0.11 & 7.48$\times$ & -0.31 & 3.85$\times$ & -1.76   \\
GHAP           & 5.02$\times$ & 0.00  & 5.06$\times$ & -0.57 & 4.87$\times$ & -0.48   \\
Mini-splatting & 4.39$\times$ & +0.27 & 5.23$\times$ & -0.33 & 6.17$\times$ & -3.48   \\
Taming-3dgs    & 8.48$\times$ & +0.09 & 4.91$\times$ & +0.04 & 5.14$\times$ & -0.12   \\
FastGS         & 7.69$\times$ & +0.30 & 6.28$\times$ & +0.30 & 4.40$\times$ & -0.02   \\
\midrule
\rowcolor{pink!40} MMGS (Ours)    & \textbf{10.25$\times$} & \textbf{+0.44} & \textbf{8.26$\times$} & \textbf{+0.31} & \textbf{8.41$\times$} & \textbf{+0.03} \\
\bottomrule
\end{tabular}
}
\vspace{-10pt}
\end{table*}

\begin{table*}[t]
\caption{\textbf{Quantitative comparison on Mip-NeRF 360, Tanks $\&$ Temples, and Deep Blending datasets.} Best results are \textbf{bolded}. The color of each cell shows the \colorbox[HTML]{FD6864}{best} and the \colorbox[HTML]{FFCC67}{second best}.}
\label{tab:main_metrics}
\center
\renewcommand{\arraystretch}{1.1}
\setlength{\tabcolsep}{1pt}
\resizebox{1.0\linewidth}{!}{
\begin{tabular}{l|ccccc|ccccc|ccccc}
\toprule
Dataset & \multicolumn{5}{c|}{Mip-NeRF 360~\cite{barron2022mip}} & \multicolumn{5}{c|}{Tanks $\&$ Temples~\cite{Knapitsch2017}} & \multicolumn{5}{c}{Deep Blending~\cite{hedman2018deep}} \\
Metric  & Time↓ & PSNR↑ & SSIM↑ & LPIPS↓ & $N_{GS}$↓ & Time↓ & PSNR↑ & SSIM↑ & LPIPS↓ & $N_{GS}$↓ & Time↓ & PSNR↑ & SSIM↑ & LPIPS↓ & $N_{GS}$↓ \\
\midrule
3DGS           & 30.46 & 29.06 & 0.870 & 0.185 & 2.47M & 17.80 & 23.76 & \cellcolor[HTML]{FFCC67}0.850 & \cellcolor[HTML]{FD6864}0.180 & 1.57M & 30.18 & 29.74 & 0.903 & 0.250 & 2.46M \\
LightGaussian  & 27.20 & 27.15 & 0.817 & 0.253 & 0.79M & 16.50 & 23.15 & 0.820 & 0.229 & 0.47M & 25.73 & 27.25 & 0.877 & 0.298 & 0.57M \\
Compact-3DGS   & 34.13 & 27.46 & 0.816 & 0.246 & 1.35M & 20.45 & 23.26 & 0.831 & 0.201 & 0.83M & 30.50 & 29.67 & 0.900 & 0.258 & 1.04M \\
RadSplat       & 27.83 & 27.54 & 0.825 & 0.239 & 0.71M & 17.44 & 23.38 & 0.831 & 0.208 & 0.42M & 28.02 & 29.98 & \cellcolor[HTML]{FD6864}0.908 & 0.255 & 0.90M \\
DashGaussian   & 10.81 & \cellcolor[HTML]{FFCC67}29.12 & \cellcolor[HTML]{FFCC67}0.872 & 0.186 & 2.22M & 7.22  & 23.94 & 0.847 & 0.181 & 1.21M & 7.18  & 29.61 & 0.902 & 0.249 & 1.94M \\
Speedy-splat   & 24.34 & 28.89 & 0.869 & 0.198 & 0.52M & 13.50 & 23.45 & 0.820 & 0.240 & \cellcolor[HTML]{FFCC67}0.21M & 18.60 & 29.63 & 0.905 & 0.256 & 0.48M \\
GHAP           & 21.33 & 28.52 & 0.856 & 0.219 & 0.49M & 12.13 & 23.19 & 0.828 & 0.217 & 0.31M & 20.26 & 29.74 & 0.902 & 0.255 & 0.49M \\
Mini-splatting & 26.07 & 29.02 & 0.869 & \cellcolor[HTML]{FD6864}0.170 & 0.47M & 16.63 & 23.43 & 0.844 & \cellcolor[HTML]{FFCC67}0.181 & 0.30M & 21.84 & 30.01 & \cellcolor[HTML]{FFCC67}0.907 & \cellcolor[HTML]{FD6864}0.243 & 0.56M \\
Taming-3dgs    & 6.78  & 28.69 & 0.852 & 0.223 & 0.66M & 4.85  & 23.80 & 0.833 & 0.212 & 0.32M & 4.05  & 29.83 & 0.899 & 0.274 & \cellcolor[HTML]{FFCC67}0.29M \\
FastGS         & \cellcolor[HTML]{FFCC67}4.15  & 28.91 & 0.864 & 0.207 & \cellcolor[HTML]{FFCC67}0.42M & \cellcolor[HTML]{FFCC67}2.84  & 24.06 & 0.835 & 0.212 & 0.25M & \cellcolor[HTML]{FD6864}2.98 & 30.04 & 0.904 & 0.257 & 0.32M \\
\midrule
\rowcolor{pink!40} 
MMGS (ours)    & \cellcolor[HTML]{FD6864}\textbf{4.08} & 28.89 & 0.861 & 0.215 & \cellcolor[HTML]{FD6864}\textbf{0.29M} & \cellcolor[HTML]{FD6864}\textbf{2.83} & \cellcolor[HTML]{FFCC67}24.07 & 0.838 & 0.208 & \cellcolor[HTML]{FD6864}\textbf{0.19M} & \cellcolor[HTML]{FFCC67}3.04 & \cellcolor[HTML]{FFCC67}30.18 & 0.905 & 0.260 & \cellcolor[HTML]{FD6864}\textbf{0.24M} \\
\rowcolor{pink!40} 
MMGS-B (ours)  & 6.11 & \cellcolor[HTML]{FD6864}\textbf{29.17} & \cellcolor[HTML]{FD6864}\textbf{0.885} & \cellcolor[HTML]{FFCC67}0.184 & 0.57M & 4.02  & \cellcolor[HTML]{FD6864}\textbf{24.22} & \cellcolor[HTML]{FD6864}\textbf{0.851} & \cellcolor[HTML]{FD6864}\textbf{0.180} & 0.34M & 4.05 & \cellcolor[HTML]{FD6864}\textbf{30.27} & \cellcolor[HTML]{FD6864}\textbf{0.908} & \cellcolor[HTML]{FFCC67}0.248 & 0.46M \\
\bottomrule 
\end{tabular}
}
\vspace{-10pt}
\end{table*}

\section{Experiments}

\subsection{Experiments Setup}

\paragraph{Datasets and Metrics.}

In accordance with prior research, we utilize four different datasets, comprising seven publicly available scenes from Mip-NeRF 360~\cite{barron2022mip}, two outdoor scenes from Tanks $\&$ Temples~\cite{Knapitsch2017}, and two indoor scenes from Deep Blending~\cite{hedman2018deep}. 
Besides, we conduct experiments on Waymo~\cite{Sun_2020_CVPRwaymo} dataset, a large-scale driving dataset with realistic scenes, which requires robustness to limited viewpoint and partial occlusions.

Consistent with previous evaluation protocols, we employ Peak Signal-to-Noise Ratio (PSNR), Structural Similarity Index Measure (SSIM)~\cite{wang2004image}, and Learned Perceptual Image Patch Similarity (LPIPS)~\cite{zhang2018unreasonable} as quantitative metrics.
We also report the training time, rendering efficiency (FPS) and final number of Gaussians.
We report the average scores for each dataset, with detailed results including GPU peak memory usage provided in the Appendix~\ref{der}.

\paragraph{Baselines.}
The benchmark model is vanilla 3DGS~\cite{kerbl20233d}.
Other methods include
\text{DashGaussian}~\cite{chen2025dashgaussian},
LightGaussian~\cite{fan2024lightgaussian},
RadSplat~\cite{niemeyer2025radsplat},
Compact3DGS~\cite{lee2024compact},
\text{Taming-3dgs}~\cite{mallick2024taming},
Mini-splatting~\cite{fang2024mini},
\text{GHAP}~\cite{wang2025gaussian},
\text{Speedy-splat}~\cite{hanson2025speedy} and
\text{FastGS}~\cite{ren2025fastgs}.

\paragraph{Implementation details.}
All experiments are conducted in PyTorch and run on one NVIDIA A6000 GPU. 
We reimplement all baselines on the same device and the same CUDA version.
We follow the standard protocols used in prior work, adopting the same training/testing splits and evaluation metrics to ensure comparability.
For GHAP, we use vanilla 3DGS as its baseline.
The sample ratio is set to 0.8 by default.
The splitting factor $\eta$ is set to 0.45.
The ranking threshold $\tau_1$ is set to 0.1 and $\tau_2$ is set to 0.05.
More detailed results and peak GPU memory report can be found in the Appendix.
For the MMGS-B, we reduce the threshold of densification process for better rendering quality with more Gaussians reserved.
Our loss function for training is the same as vanilla 3DGS.

\begin{figure*}[t]
\vspace{-15pt}
  \begin{center}
    \centerline{\includegraphics[width=1.0\columnwidth, trim=5.6cm 6.5cm 0cm 0cm, clip]{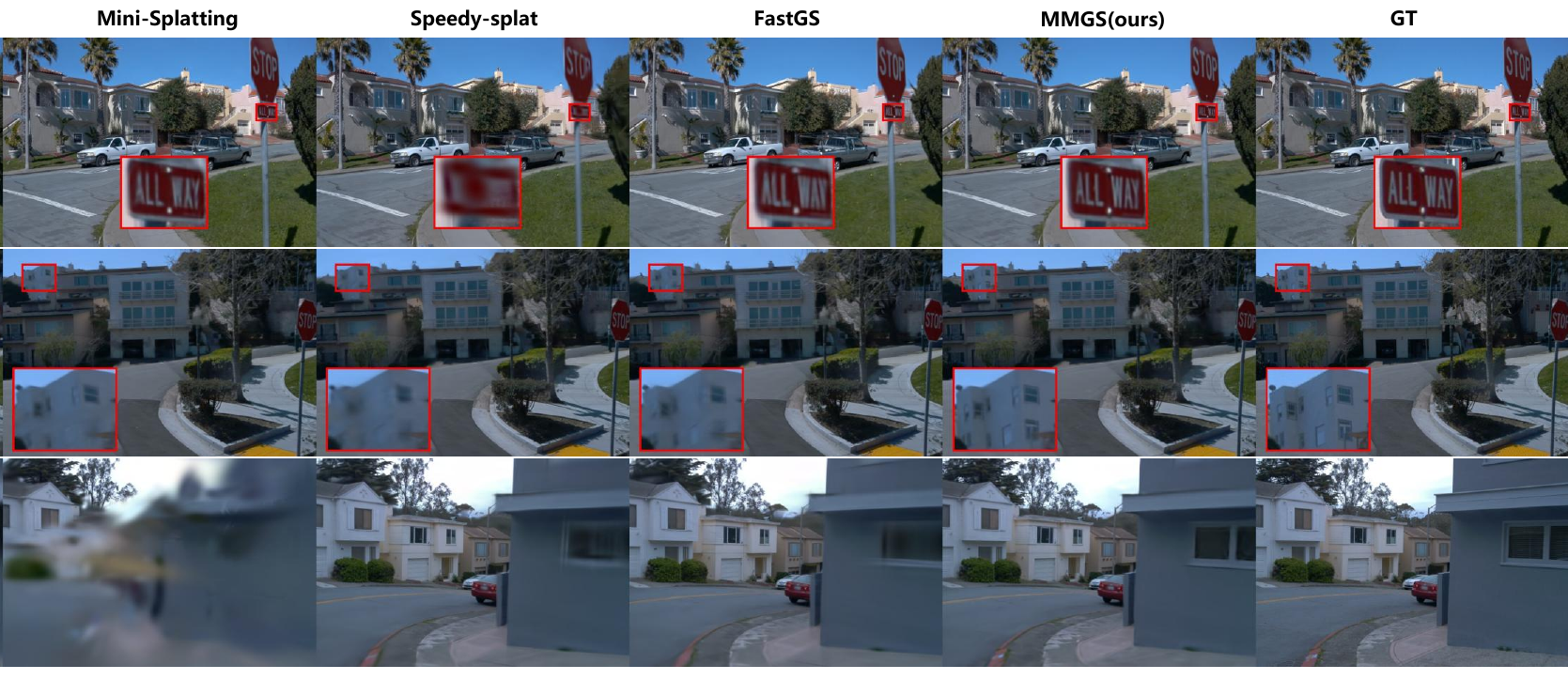}}
     \vspace{-9pt}
    \caption{\textbf{Comparison Results on Waymo~\cite{Sun_2020_CVPRwaymo} dataset.} Visual differences are highlighted with red insets for better clarity. Our approach consistently outperforms other models on different scenes, demonstrating advantages in challenging scenarios. Best viewed in color.}
    \label{e1}
  \end{center}
  \vspace{-20pt}
\end{figure*}

\begin{table*}[t]
\small
\caption{\textbf{Quantitative comparison on Waymo~\cite{Sun_2020_CVPRwaymo} dataset.} This dataset tests the ability of methods to fit in large scale outdoor environments. The color of each cell shows the \colorbox[HTML]{FD6864}{best} and the \colorbox[HTML]{FFCC67}{second best}.}
\label{tab:waymo_metrics}
\center
\setlength{\tabcolsep}{6pt}
\begin{tabular}{l|cccccc}
\toprule
Metric  & Time↓ & PSNR↑ & SSIM↑ & LPIPS↓ & $N_{GS}$↓ & FPS↑  \\
\midrule
Speedy-splat   & 31.54 & 25.14 & 0.805 & 0.363 & 0.48M & 216 \\
GHAP           & 31.21 & 26.42 & 0.824 & 0.345 & 0.38M & \cellcolor[HTML]{FFCC67}361 \\
Mini-splatting & 43.44 & 23.42 & 0.784 & 0.391 & \cellcolor[HTML]{FFCC67}0.30M & 360 \\
Taming-3dgs    & 10.18 & 26.78 & 0.824 & 0.351 & 0.36M & 146 \\
FastGS         & \cellcolor[HTML]{FFCC67}5.67  & 26.88 & 0.826 & 0.349 & 0.42M & 283 \\
\midrule 
\rowcolor{pink!40} 
MMGS (ours)    & \cellcolor[HTML]{FD6864}\textbf{4.36} & \cellcolor[HTML]{FFCC67}26.93 & \cellcolor[HTML]{FFCC67}0.828 & \cellcolor[HTML]{FFCC67}0.344 & \cellcolor[HTML]{FD6864}\textbf{0.22M} & \cellcolor[HTML]{FD6864}\textbf{367} \\
\rowcolor{pink!40} 
MMGS-B (ours)  & 6.55  & \cellcolor[HTML]{FD6864}\textbf{27.24} & \cellcolor[HTML]{FD6864}\textbf{0.833} & \cellcolor[HTML]{FD6864}\textbf{0.331} & 0.52M & 303 \\
\bottomrule 
\end{tabular}
\vspace{-15pt}
\end{table*}

\subsection{Quantitative Results}

The quantitative comparisons are summarized in Tab.~\ref{tab:ratio_comparison}, Tab.~\ref{tab:main_metrics} and Tab.~\ref{tab:waymo_metrics}. 
We visualize rendered results in Fig.~\ref{e1} and Fig.~\ref{e2}.
More detailed results such as FPS and GPU peak memory are reported in the Appendix.~\ref{detailed results}.

\paragraph{Rendering Quality vs. Primitive Count.} 
Compared to vanilla 3DGS, our method can speed up the training process by 10$×$ with 10$\%$ Gaussians while exceeding metrics scores. 
While pruning-based methods like Mini-Splatting and GHAP successfully reduce primitive counts, they often suffer from a noticeable drop in PSNR due to the loss of high-frequency texture details. 
In contrast, our Optimal Transport-based aggregation ensures that the representational capacity of the reduced set mathematically approximates the original distribution, preserving visual fidelity.

\paragraph{Efficiency and Robustness.}
In terms of training speed, our method converges significantly faster than Taming-3DGS and DashGaussian due to the reduced computational graph size during the densification stages. 
On Waymo dataset, which suffers from sparse supervision, standard gradient-based densification tends to generate floaters near the camera. 
Our multi-view 3D ranking strategy effectively suppresses these artifacts, resulting in cleaner geometry and higher evaluation metrics in unbounded driving scenes.

\paragraph{Runtime of Each Module.}

Regarding the runtime overhead of the OT aggregation step, thanks to our highly optimized Weighted KD-Tree and batched Wasserstein feature-mapping, the global OT aggregation is highly efficient. 
As detailed in Tab.~\ref{tab:time_breakdown}, a single execution of the global OT aggregation step takes only about \textbf{10 seconds}. 
This transient computational overhead is instantly amortized by the drastically reduced computational graph size for all subsequent training iterations.

\begin{figure*}[t]
\vspace{-10pt}
  \begin{center}
    \centerline{\includegraphics[width=1.0\columnwidth, trim=5.6cm 6.9cm 0cm 0cm, clip]{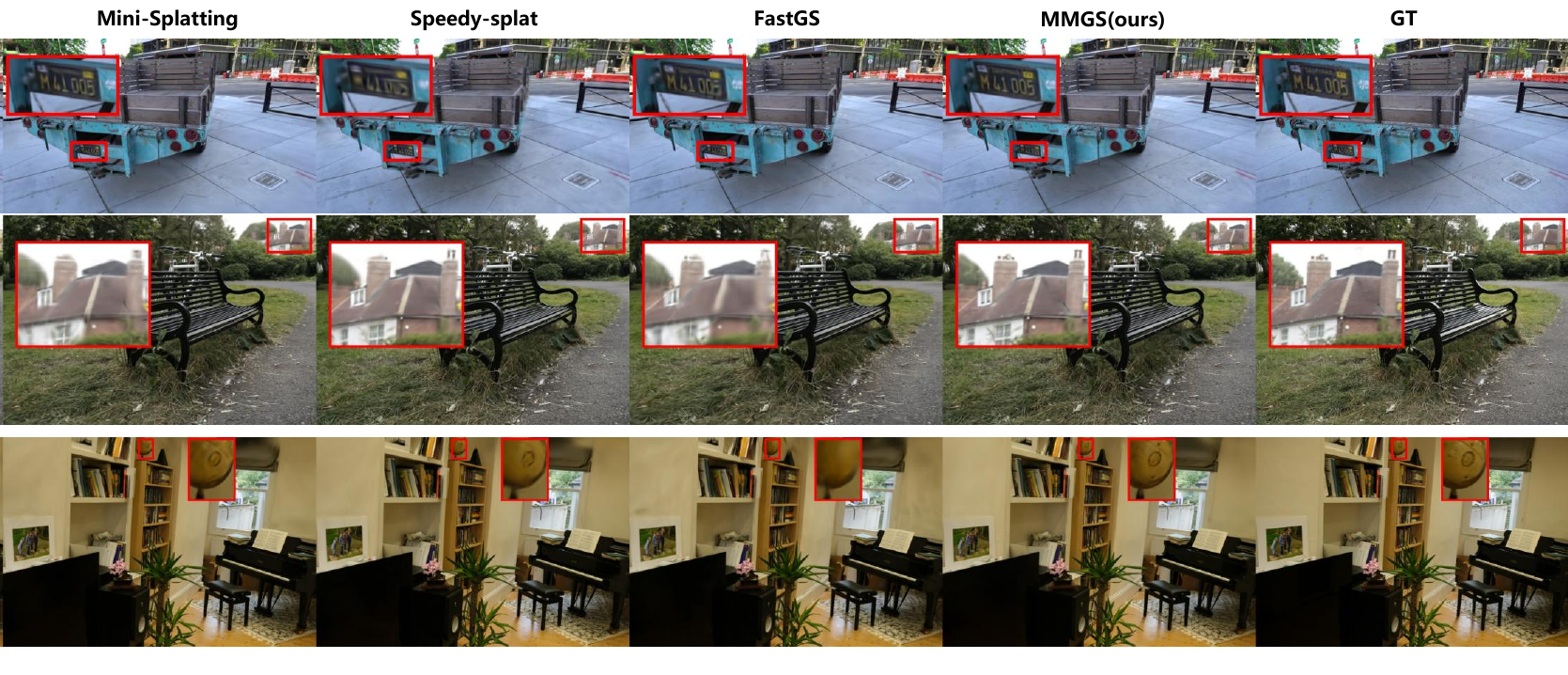}}
     \vspace{-9pt}
    \caption{\textbf{Comparison Results on Mip-NeRF360~\cite{barron2022mip}, Tanks $\&$ Temples~\cite{Knapitsch2017} Deep Blending~\cite{hedman2018deep} dataset.} Visual differences are highlighted with red insets for better clarity. Our approach consistently outperforms other models on different scenes, demonstrating advantages in challenging scenarios. Best viewed in color.}
    \label{e2}
  \end{center}
  \vspace{-20pt}
\end{figure*}

\begin{table}[t]
\centering
\caption{\textbf{Detailed runtime of each module on Mip-NeRF360 dataset.}}
\label{tab:time_breakdown}
\resizebox{\textwidth}{!}{ 
\begin{tabular}{lccccccc}
\toprule
 & bicycle & bonsai & counter & garden & kitchen & room & stump \\
\midrule
Total Training time & 4 min 17 s & 4 min 12 s & 4 min 18 s & 3 min 55 s & 4 min 47 s & 3 min 47 s & 3 min 33 s \\
Time for Ranking & 10.75 s & 12.46 s & 12.58 s & 10.91 s & 13.16 s & 11.71 s & 10.14 s \\
Time for OT Agg. & 13.92 s & 8.22 s & 7.67 s & 10.74 s & 8.55 s & 7.80 s & 12.43 s \\
\bottomrule
\end{tabular}
}
\vspace{-10pt}
\end{table}

\subsection{Ablation Study}

To validate the effectiveness of our design choices, we conduct a comprehensive ablation study on Mip-NeRF 360 dataset.
The ablation study is divided into two parts. 
First, we conduct standard ablation experiments for each component to validate their effectiveness.
Then we discuss the design choice within individual components.
The results are shown in Tab.~\ref{ab1} and Tab.~\ref{ab2}.

\subsubsection{Component-wise Analysis}
We remove our three core components individually to the vanilla 3DGS baseline. 

\paragraph{Effectiveness of Multi-View 3D Ranking.} 
Replacing the standard gradient heuristic with our multi-view 3D ranking significantly improves geometry and reduces the number of Gaussians. 
By filtering out primitives that are not geometrically consistent across views, we eliminate a large number of meaningless artifacts. 
This manifests quantitatively as an improvement in visual metrics and fewer primitives.

\paragraph{Effectiveness of OT-Based Aggregation.} 
The inclusion of Optimal Transport aggregation provides the most significant reduction in primitive count. 
Unlike simple distance-based merging, OT aggregation considers the transport cost of radiance, allowing us to merge large clusters of redundant Gaussians into a single, optimized primitive without losing structural coverage. 
This module is responsible for the massive memory savings with negligible quality loss.
    
\paragraph{Effectiveness of Physically Grounded Densification.} Finally, enabling the Optimal Transport splitting strategies stabilizes the optimization. 
This prevents the popping issues in heuristic splitting and allows the model to converge to a lower loss value, further boosting visual quality.

\begin{table*}[ht] 
\vspace{-13pt}
\center
\begin{minipage}{0.48\textwidth}
\center
\small
\caption{\textbf{Component-wise Ablation Analysis on Mip-NeRF 360~\cite{barron2022mip} dataset.}}
\label{ab1}
\setlength{\tabcolsep}{2pt} 
\resizebox{\linewidth}{!}{
\begin{tabular}{lcccccc}
\toprule
\multicolumn{1}{c}{Metric} & Time↓  & PSNR↑   & SSIM↑  & LPIPS↓ & $N_{GS}$↓  & FPS↑ \\
\midrule
w/o Ranking                & 4.19 & 28.83 & 0.859 & 0.217 & 0.41M & 309 \\
w/o OT Agg.                & 4.11 & 28.76 & 0.858 & 0.216 & 0.35M & 321 \\
w/o OT Split               & 4.12 & 28.62 & 0.851 & 0.229 & 0.32M & 344 \\
\midrule
Ours     & \textbf{4.08} & \textbf{28.87} & \textbf{0.861} & \textbf{0.215} & \textbf{0.29M} & \textbf{355} \\
\bottomrule
\end{tabular}
}
\end{minipage}
\hfill
\begin{minipage}{0.48\textwidth}
\center
\small
\caption{\textbf{Micro-Ablation of Design Choices on Mip-NeRF 360~\cite{barron2022mip} dataset.}}
\label{ab2}
\setlength{\tabcolsep}{2pt}
\resizebox{\linewidth}{!}{
\begin{tabular}{lcccccc}
\toprule
\multicolumn{1}{c}{Metric} & Time↓  & PSNR↑   & SSIM↑  & LPIPS↓ & $N_{GS}$↓  & FPS↑  \\
\midrule
w/ GHAP                     & 4.09 & 28.71 & 0.856 & 0.226 & 0.32M & 333 \\
w/ L1 ONLY                  & 4.17 & 28.85 & 0.861 & 0.218 & 0.41M & 326 \\
w/ KD-Tree                  & 4.15 & 28.55 & 0.855 & 0.244 & 0.31M & 305 \\
\midrule
Ours     & \textbf{4.08} & \textbf{28.87} & \textbf{0.861} & \textbf{0.215} & \textbf{0.29M} & \textbf{355} \\
\bottomrule
\end{tabular}
}
\end{minipage}
\vspace{-10pt}
\end{table*}

\subsubsection{Micro-Ablation of Design Choices}
We further analyze specific sub-modules to justify our architectural decisions.
The computational complexity analysis and micro-ablation of our importance-balanced KD-Tree algorithm are detailed in Appendix.~\ref{supp:complexity}.

\paragraph{Importance-balanced KD-Tree vs. KD-Tree.} 
Standard KD-Trees partition the point cloud based on median coordinates, which ignores the highly non-uniform information density of radiance fields. 
In contrast, our algorithm uses the visual contribution score $\mathcal{W}_i$ to ensure that the partitioned blocks are informationally equal. 
The results indicates that incorporating visual contribution mass into the KD-Tree construction is crucial for preserving fine details effectively.

\paragraph{Texture-Aware Metric vs. L1 Loss.} We replace our texture-aware structural error metric with a standard RGB L1 loss. 
The results indicate that the texture-aware metric is crucial for identifying high-frequency regions. 
Without it, the model fails to prioritize complex textures during densification, leading to blurrier results at the same Gaussian budget or more meaningless primitives.

\begin{wraptable}{r}{0.5\textwidth}
  \vspace{-10pt}
  \begin{center}
    \centerline{\includegraphics[width=0.5\columnwidth]{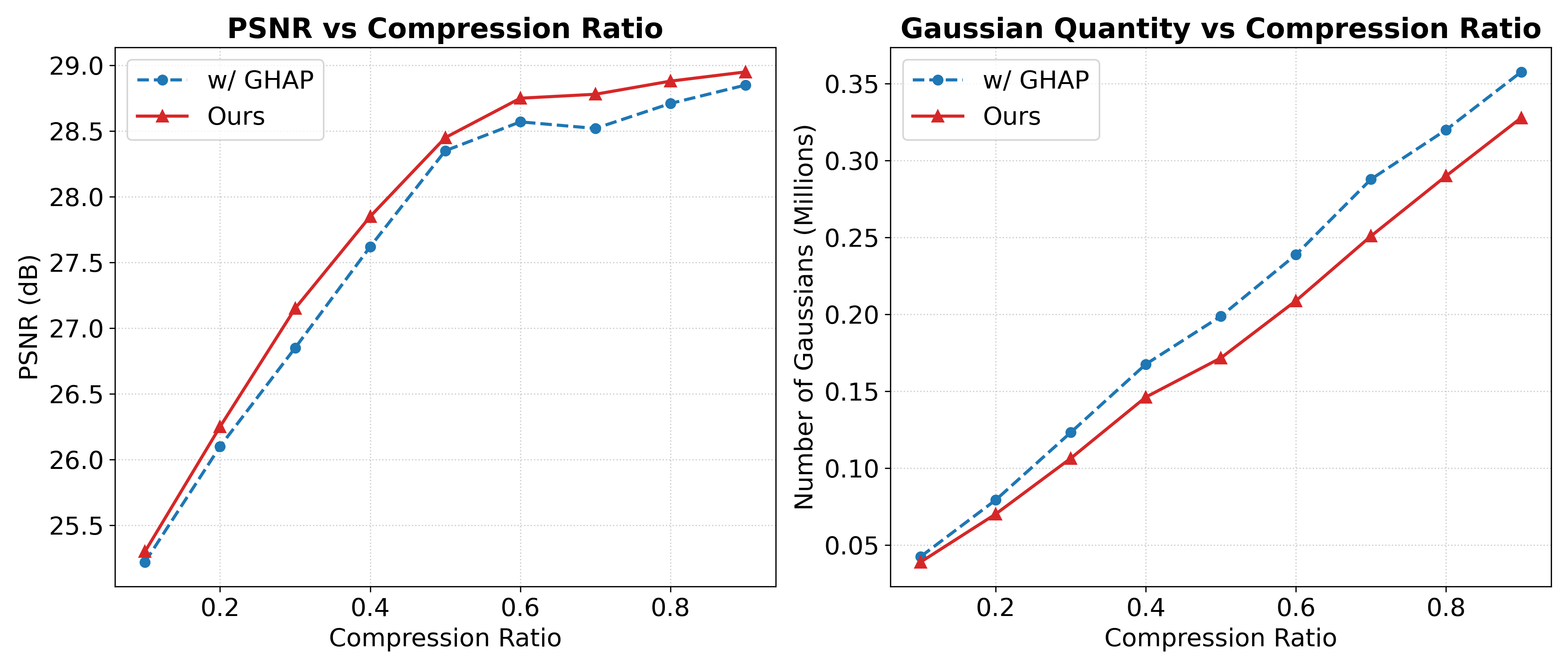}}
    \caption{\textbf{Ablation Results.} Quantitative results of different sample ratio of OT aggregation algorithm and the comparison with GHAP. Our algorithm can reach consistently better PSNR while maintain less primitives.}
    \label{ablation}
  \end{center}
   \vspace{-30pt}
\end{wraptable}

\paragraph{OT Aggregation vs. GHAP.} 
As shown in Fig.~\ref{ablation}, we compare our OT aggregation algorithm against GHAP~\cite{wang2025gaussian}. 
Based on their implementation, the cost matrix is computed using a direct Euclidean distance and the Frobenius norm of the simple difference between covariance matrices: $C_{ij} = ||\mu_1 - \mu_2||_2^2 + ||\Sigma_1 - \Sigma_2||_F^2$. 
Computing direct linear subtraction ignores non-Euclidean structure and contradicts with OT theory. 
Our OT formulation solves a global distribution matching problem, which better preserves the important Gaussians of the scene. 
Consequently, at same sample ratio, our method outperforms GHAP by a significant margin.

\subsection{Discussion and Limitation}

Optimization-based Gaussian splatting relies heavily on 2D view-space gradients to infer 3D structure. 
Our analysis suggests that standard 3DGS conflates these errors, leading to geometric degradation and excessive proliferation of primitives to overfit 2D projections. 
By introducing explicit 3D spatial constraints and multi-view consistency, we mitigate this inherent flaw, proving that accurate 3D priors are essential for efficient optimization.
A novel optimization and rendering algorithm designed for 3DGS is required to completely resolve this issue.

\section{Conclusion}

In this work, we addressed the inefficiencies in 3D Gaussian Splatting arising from heuristic 2D gradient supervision. 
We proposed MMGS, a novel framework which redefines primitive optimization as a 3D distribution matching problem grounded in Optimal Transport theory based on contribution ranking. 
By enforcing multi-view geometric consistency, we decoupled primitive generation from ambiguous view-dependent projection errors, thereby eliminating floating artifacts and ensuring structural integrity. 
Furthermore, our novel OT-based aggregation and densification strategies enabled the reduction of redundancy, compressing the scene representation by \textbf{90$\%$} while maintaining state-of-the-art visual fidelity and real-time rendering.

\section{Acknowledgment}
This work is supported by the National Natural Science Foundation of China (No. 62406267), Guangdong Provincial Project (No. 2024QN11X072), Guangzhou-HKUST(GZ) Joint Funding Program (No. 2025A03J3956) and Guangzhou Municipal Education Project (No. 2024312122).

\bibliographystyle{unsrt}
\bibliography{example_paper}

@article{fei20243d,
  title={3d gaussian splatting as new era: A survey},
  author={Fei, Ben and Xu, Jingyi and Zhang, Rui and Zhou, Qingyuan and Yang, Weidong and He, Ying},
  journal={IEEE Transactions on Visualization and Computer Graphics},
  year={2024},
  publisher={IEEE}
}

@article{wu2024recent,
  title={Recent advances in 3d gaussian splatting},
  author={Wu, Tong and Yuan, Yu-Jie and Zhang, Ling-Xiao and Yang, Jie and Cao, Yan-Pei and Yan, Ling-Qi and Gao, Lin},
  journal={Computational Visual Media},
  volume={10},
  number={4},
  pages={613--642},
  year={2024},
  publisher={TUP}
}

@article{bao20253d,
  title={3d gaussian splatting: Survey, technologies, challenges, and opportunities},
  author={Bao, Yanqi and Ding, Tianyu and Huo, Jing and Liu, Yaoli and Li, Yuxin and Li, Wenbin and Gao, Yang and Luo, Jiebo},
  journal={IEEE Transactions on Circuits and Systems for Video Technology},
  year={2025},
  publisher={IEEE}
}

@article{ren2024octree,
  title={Octree-gs: Towards consistent real-time rendering with lod-structured 3d gaussians},
  author={Ren, Kerui and Jiang, Lihan and Lu, Tao and Yu, Mulin and Xu, Linning and Ni, Zhangkai and Dai, Bo},
  journal={arXiv preprint arXiv:2403.17898},
  year={2024}
}

@inproceedings{lu2024scaffold,
  title={Scaffold-gs: Structured 3d gaussians for view-adaptive rendering},
  author={Lu, Tao and Yu, Mulin and Xu, Linning and Xiangli, Yuanbo and Wang, Limin and Lin, Dahua and Dai, Bo},
  booktitle={Proceedings of the IEEE/CVF Conference on Computer Vision and Pattern Recognition},
  pages={20654--20664},
  year={2024}
}

@inproceedings{bagdasarian20253dgs,
  title={3dgs. zip: A survey on 3d gaussian splatting compression methods},
  author={Bagdasarian, Milena T and Knoll, Paul and Li, Y and Barthel, Florian and Hilsmann, Anna and Eisert, Peter and Morgenstern, Wieland},
  booktitle={Computer Graphics Forum},
  volume={44},
  number={2},
  pages={e70078},
  year={2025},
  organization={Wiley Online Library}
}

@article{liu2024atomgs,
  title={Atomgs: Atomizing gaussian splatting for high-fidelity radiance field},
  author={Liu, Rong and Xu, Rui and Hu, Yue and Chen, Meida and Feng, Andrew},
  journal={arXiv preprint arXiv:2405.12369},
  year={2024}
}

@article{ali2024trimming,
  title={Trimming the fat: Efficient compression of 3d gaussian splats through pruning},
  author={Ali, Muhammad Salman and Qamar, Maryam and Bae, Sung-Ho and Tartaglione, Enzo},
  journal={arXiv preprint arXiv:2406.18214},
  year={2024}
}

@inproceedings{chen2024hac,
  title={Hac: Hash-grid assisted context for 3d gaussian splatting compression},
  author={Chen, Yihang and Wu, Qianyi and Lin, Weiyao and Harandi, Mehrtash and Cai, Jianfei},
  booktitle={European Conference on Computer Vision},
  pages={422--438},
  year={2024},
  organization={Springer}
}

@book{villani2008optimal,
  title={Optimal transport: old and new},
  author={Villani, C{\'e}dric and others},
  volume={338},
  year={2008},
  publisher={Springer}
}

@article{thorpe2018introduction,
  title={Introduction to optimal transport},
  author={Thorpe, Matthew},
  journal={Notes of Course at University of Cambridge},
  volume={3},
  year={2018}
}

@article{montesuma2024recent,
  title={Recent advances in optimal transport for machine learning},
  author={Montesuma, Eduardo Fernandes and Mboula, Fred Maurice Ngole and Souloumiac, Antoine},
  journal={IEEE Transactions on Pattern Analysis and Machine Intelligence},
  year={2024},
  publisher={IEEE}
}

@article{moon1996expectation,
  title={The expectation-maximization algorithm},
  author={Moon, Todd K},
  journal={IEEE Signal processing magazine},
  volume={13},
  number={6},
  pages={47--60},
  year={1996},
  publisher={IEEE}
}

@article{ruschendorf1985wasserstein,
  title={The Wasserstein distance and approximation theorems},
  author={R{\"u}schendorf, Ludger},
  journal={Probability Theory and Related Fields},
  volume={70},
  number={1},
  pages={117--129},
  year={1985},
  publisher={Springer}
}

@article{bhatia2019bures,
  title={On the Bures--Wasserstein distance between positive definite matrices},
  author={Bhatia, Rajendra and Jain, Tanvi and Lim, Yongdo},
  journal={Expositiones mathematicae},
  volume={37},
  number={2},
  pages={165--191},
  year={2019},
  publisher={Elsevier}
}

@article{gelbrich1990formula,
  title={On a formula for the L2 Wasserstein metric between measures on Euclidean and Hilbert spaces},
  author={Gelbrich, Matthias},
  journal={Mathematische Nachrichten},
  volume={147},
  number={1},
  pages={185--203},
  year={1990},
  publisher={Wiley Online Library}
}

@article{ren2025fastgs,
  title={FastGS: Training 3D Gaussian Splatting in 100 Seconds},
  author={Ren, Shiwei and Wen, Tianci and Fang, Yongchun and Lu, Biao},
  journal={arXiv preprint arXiv:2511.04283},
  year={2025}
}

@article{kerbl20233d,
  title={3D Gaussian splatting for real-time radiance field rendering.},
  author={Kerbl, Bernhard and Kopanas, Georgios and Leimk{\"u}hler, Thomas and Drettakis, George},
  journal={ACM Trans. Graph.},
  volume={42},
  number={4},
  pages={139--1},
  year={2023}
}

@inproceedings{mallick2024taming,
  title={Taming 3dgs: High-quality radiance fields with limited resources},
  author={Mallick, Saswat Subhajyoti and Goel, Rahul and Kerbl, Bernhard and Steinberger, Markus and Carrasco, Francisco Vicente and De La Torre, Fernando},
  booktitle={SIGGRAPH Asia 2024 Conference Papers},
  pages={1--11},
  year={2024}
}

@article{wang2025gaussian,
  title={Gaussian Herding across Pens: An Optimal Transport Perspective on Global Gaussian Reduction for 3DGS},
  author={Wang, Tao and Li, Mengyu and Zeng, Geduo and Meng, Cheng and Zhang, Qiong},
  journal={arXiv preprint arXiv:2506.09534},
  year={2025}
}

@inproceedings{fang2024mini,
  title={Mini-splatting: Representing scenes with a constrained number of gaussians},
  author={Fang, Guangchi and Wang, Bing},
  booktitle={European Conference on Computer Vision},
  pages={165--181},
  year={2024},
  organization={Springer}
}

@inproceedings{chen2025dashgaussian,
  title={DashGaussian: Optimizing 3D Gaussian Splatting in 200 Seconds},
  author={Chen, Youyu and Jiang, Junjun and Jiang, Kui and Tang, Xiao and Li, Zhihao and Liu, Xianming and Nie, Yinyu},
  booktitle={Proceedings of the Computer Vision and Pattern Recognition Conference},
  pages={11146--11155},
  year={2025}
}

@inproceedings{hanson2025speedy,
  title={Speedy-splat: Fast 3d gaussian splatting with sparse pixels and sparse primitives},
  author={Hanson, Alex and Tu, Allen and Lin, Geng and Singla, Vasu and Zwicker, Matthias and Goldstein, Tom},
  booktitle={Proceedings of the Computer Vision and Pattern Recognition Conference},
  pages={21537--21546},
  year={2025}
}

@article{radl2024stopthepop,
  title={Stopthepop: Sorted gaussian splatting for view-consistent real-time rendering},
  author={Radl, Lukas and Steiner, Michael and Parger, Mathias and Weinrauch, Alexander and Kerbl, Bernhard and Steinberger, Markus},
  journal={ACM Transactions on Graphics (TOG)},
  volume={43},
  number={4},
  pages={1--17},
  year={2024},
  publisher={ACM New York, NY, USA}
}

@inproceedings{feng2025flashgs,
  title={Flashgs: Efficient 3d gaussian splatting for large-scale and high-resolution rendering},
  author={Feng, Guofeng and Chen, Siyan and Fu, Rong and Liao, Zimu and Wang, Yi and Liu, Tao and Hu, Boni and Xu, Linning and Pei, Zhilin and Li, Hengjie and others},
  booktitle={Proceedings of the Computer Vision and Pattern Recognition Conference},
  pages={26652--26662},
  year={2025}
}

@article{kheradmand20243d,
  title={3d gaussian splatting as markov chain monte carlo},
  author={Kheradmand, Shakiba and Rebain, Daniel and Sharma, Gopal and Sun, Weiwei and Tseng, Yang-Che and Isack, Hossam and Kar, Abhishek and Tagliasacchi, Andrea and Yi, Kwang Moo},
  journal={Advances in Neural Information Processing Systems},
  volume={37},
  pages={80965--80986},
  year={2024}
}

@inproceedings{cheng2024gaussianpro,
  title={Gaussianpro: 3d gaussian splatting with progressive propagation},
  author={Cheng, Kai and Long, Xiaoxiao and Yang, Kaizhi and Yao, Yao and Yin, Wei and Ma, Yuexin and Wang, Wenping and Chen, Xuejin},
  booktitle={Forty-first International Conference on Machine Learning},
  year={2024}
}

@article{fan2024lightgaussian,
  title={Lightgaussian: Unbounded 3d gaussian compression with 15x reduction and 200+ fps},
  author={Fan, Zhiwen and Wang, Kevin and Wen, Kairun and Zhu, Zehao and Xu, Dejia and Wang, Zhangyang and others},
  journal={Advances in neural information processing systems},
  volume={37},
  pages={140138--140158},
  year={2024}
}

@inproceedings{niemeyer2025radsplat,
  title={Radsplat: Radiance field-informed gaussian splatting for robust real-time rendering with 900+ fps},
  author={Niemeyer, Michael and Manhardt, Fabian and Rakotosaona, Marie-Julie and Oechsle, Michael and Duckworth, Daniel and Gosula, Rama and Tateno, Keisuke and Bates, John and Kaeser, Dominik and Tombari, Federico},
  booktitle={2025 International Conference on 3D Vision (3DV)},
  pages={134--144},
  year={2025},
  organization={IEEE}
}

@article{lee2024compact,
  title={Compact 3d gaussian splatting for static and dynamic radiance fields},
  author={Lee, Joo Chan and Rho, Daniel and Sun, Xiangyu and Ko, Jong Hwan and Park, Eunbyung},
  journal={arXiv preprint arXiv:2408.03822},
  year={2024}
}

@inproceedings{hanson2025pup,
  title={Pup 3d-gs: Principled uncertainty pruning for 3d gaussian splatting},
  author={Hanson, Alex and Tu, Allen and Singla, Vasu and Jayawardhana, Mayuka and Zwicker, Matthias and Goldstein, Tom},
  booktitle={Proceedings of the Computer Vision and Pattern Recognition Conference},
  pages={5949--5958},
  year={2025}
}

@article{zhang2024lp,
  title={Lp-3dgs: Learning to prune 3d gaussian splatting},
  author={Zhang, Zhaoliang and Song, Tianchen and Lee, Yongjae and Yang, Li and Peng, Cheng and Chellappa, Rama and Fan, Deliang},
  journal={Advances in Neural Information Processing Systems},
  volume={37},
  pages={122434--122457},
  year={2024}
}

@inproceedings{hollein20253dgs,
  title={3dgs-lm: Faster gaussian-splatting optimization with levenberg-marquardt},
  author={H{\"o}llein, Lukas and Bo{\v{z}}i{\v{c}}, Alja{\v{z}} and Zollh{\"o}fer, Michael and Nie{\ss}ner, Matthias},
  booktitle={Proceedings of the IEEE/CVF International Conference on Computer Vision},
  pages={26740--26750},
  year={2025}
}

@inproceedings{lan20253dgs2,
  title={3dgs2: Near second-order converging 3d gaussian splatting},
  author={Lan, Lei and Shao, Tianjia and Lu, Zixuan and Zhang, Yu and Jiang, Chenfanfu and Yang, Yin},
  booktitle={Proceedings of the Special Interest Group on Computer Graphics and Interactive Techniques Conference Conference Papers},
  pages={1--10},
  year={2025}
}

@InProceedings{Sun_2020_CVPRwaymo, author = {Sun, Pei and Kretzschmar, Henrik and Dotiwalla, Xerxes and Chouard, Aurelien and Patnaik, Vijaysai and Tsui, Paul and Guo, James and Zhou, Yin and Chai, Yuning and Caine, Benjamin and Vasudevan, Vijay and Han, Wei and Ngiam, Jiquan and Zhao, Hang and Timofeev, Aleksei and Ettinger, Scott and Krivokon, Maxim and Gao, Amy and Joshi, Aditya and Zhang, Yu and Shlens, Jonathon and Chen, Zhifeng and Anguelov, Dragomir}, title = {Scalability in Perception for Autonomous Driving: Waymo Open Dataset}, booktitle = {Proceedings of the IEEE/CVF Conference on Computer Vision and Pattern Recognition (CVPR)}, month = {June}, year = {2020} }

@inproceedings{barron2022mip,
  title={Mip-nerf 360: Unbounded anti-aliased neural radiance fields},
  author={Barron, Jonathan T and Mildenhall, Ben and Verbin, Dor and Srinivasan, Pratul P and Hedman, Peter},
  booktitle={Proceedings of the IEEE/CVF conference on computer vision and pattern recognition},
  pages={5470--5479},
  year={2022}
}

@article{hedman2018deep,
  title={Deep blending for free-viewpoint image-based rendering},
  author={Hedman, Peter and Philip, Julien and Price, True and Frahm, Jan-Michael and Drettakis, George and Brostow, Gabriel},
  journal={ACM Transactions on Graphics (ToG)},
  volume={37},
  number={6},
  pages={1--15},
  year={2018},
  publisher={ACM New York, NY, USA}
}

@article{Knapitsch2017,
    author    = {Arno Knapitsch and Jaesik Park and Qian-Yi Zhou and Vladlen Koltun},
    title     = {Tanks and Temples: Benchmarking Large-Scale Scene Reconstruction},
    journal   = {ACM Transactions on Graphics},
    volume    = {36},
    number    = {4},
    year      = {2017},
}

@article{wang2004image,
  title={Image quality assessment: from error visibility to structural similarity},
  author={Wang, Zhou and Bovik, Alan C and Sheikh, Hamid R and Simoncelli, Eero P},
  journal={IEEE transactions on image processing},
  volume={13},
  number={4},
  pages={600--612},
  year={2004},
  publisher={IEEE}
}

@inproceedings{zhang2018unreasonable,
  title={The unreasonable effectiveness of deep features as a perceptual metric},
  author={Zhang, Richard and Isola, Phillip and Efros, Alexei A and Shechtman, Eli and Wang, Oliver},
  booktitle={Proceedings of the IEEE conference on computer vision and pattern recognition},
  pages={586--595},
  year={2018}
}


\appendix

\section{LLM Declaration and Impact Statements}
The core method development in this research does not involve LLMs as any important, original, or non-standard components.
This paper presents work whose goal is to advance the field of 3D reconstruction and 3DGS. There are many potential societal consequences of our work, none of which we feel must be specifically highlighted here.

\section{Computational Complexity Analysis}
\label{supp:complexity}

A naive global Optimal Transport (OT) approach between $N$ source Gaussians and $K$ target Gaussians requires computing a cost matrix of size $N \times K$. 
Since $N$ typically exceeds millions, a direct solution with $O(N \cdot K)$ complexity is computationally intractable. 
Our proposed importance-balanced KD-Tree strategy can effectively reduces this complexity.

\subsection{Importance-Balanced KD-Tree Construction}
\begin{figure}[ht]
  \begin{center}
    \centerline{\includegraphics[width=0.5\columnwidth]{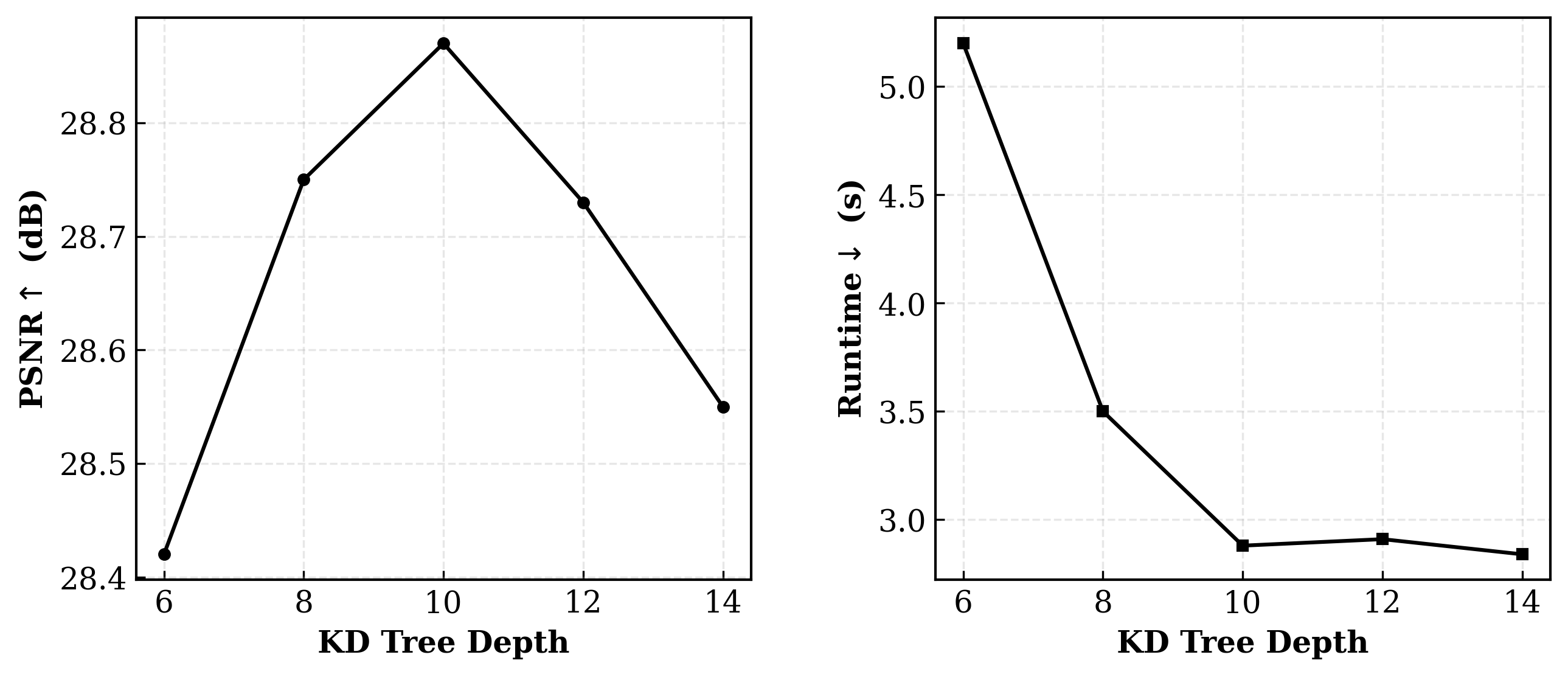}}
    \caption{\textbf{Ablation study on $D$.} The influence of different KD-Tree depth on PSNR and the runtime of the aggregation algorithm.}
    \label{kd}
  \end{center}
\end{figure}

The construction of our weighted KD-Tree involves recursive partitioning based on the cumulative distribution of the importance metric $\mathcal{W}$. 
At each depth level $d$, sorting the primitives along the cutting axis requires $O(N \log N)$. Given a tree depth $D$ (where the number of leaf nodes $M = 2^D$), the total construction complexity is:
\begin{equation}
    \mathcal{C}_{\text{tree}} = O(D \cdot N \log N).
\end{equation}
Since $D$ is a small constant (typically $10$), this step is extremely efficient and negligible compared to the rendering time.
The ablation study on $D$ is shown in Fig.~\ref{kd}.

\begin{table}[hb]
\small
\caption{\textbf{Ablation Study of Spatial Partitioning Strategies on Mip-NeRF 360 dataset.}}
\setlength{\tabcolsep}{3pt} 
\label{tab:ablation_kdtree}
\center
\sc
\begin{tabular}{lcccccc}
\toprule
\multicolumn{1}{c}{Method} & Time (s)↓ & PSNR↑ & SSIM↑ & LPIPS↓ & $N_{GS}$↓ \\
\midrule
Random & \textbf{0.52} & 26.14 & 0.792 & 0.315 & 0.36M  \\
KD-Tree & 3.95 & 28.55 & 0.855 & 0.244 & 0.31M  \\
\midrule
Ours & 4.06 & \textbf{28.87} & \textbf{0.861} & \textbf{0.215} & \textbf{0.29M}\\
\bottomrule
\end{tabular}
\end{table}

\subsection{Block-wise OT Aggregation}
Let $M$ denote the number of leaf nodes (blocks) generated by the KD-Tree. The global aggregation problem is decomposed into $M$ independent sub-problems.
Assuming the importance distribution is relatively smooth, each block contains approximately $n \approx N/M$ source Gaussians and is assigned a budget of $k \approx K/M$ target Gaussians.
The complexity of the EM-based OT solver within one block is dominated by the cost matrix computation and matrix-vector multiplications, which is $O(T \cdot n \cdot k)$, where $T$ is the number of EM iterations.

Summing over all $M$ blocks, the total complexity becomes:
\begin{equation}
    \mathcal{C}_{\text{agg}} = \sum_{i=1}^{M} O(T \cdot n_i \cdot k_i) \approx  O\left(T \cdot \frac{NK}{M}\right).
\end{equation}

By choosing an appropriate $M$, we reduce the complexity by a factor of $M$ compared to the global approach. 
In our implementation, with $N \approx 10^6$ and $M \approx 1024$, this linear scaling allows the entire reduction process to complete within a few seconds.
Finally, the reconstruction of covariance matrices via eigendecomposition is performed only on the reduced set $K$, with complexity $O(K)$, which is trivial.

\section{Derivation of OT-based Densification}
\label{supp:densification_proof}

In this section, we provide detailed mathematical proof that our densification strategy minimizes the structural divergence between the parent Gaussian and sub-Gaussian, ensuring seamless optimization transitions.

Let the original Gaussian be $g \sim \mathcal{N}(\mu, \Sigma)$. We aim to split $g$ into a mixture of two equally weighted sub-Gaussians $g_1, g_2$ with parameters $(\mu_1, \Sigma')$ and $(\mu_2, \Sigma')$.
The mixture distribution is $p_{mix}(x) = 0.5 \mathcal{N}(x; \mu_1, \Sigma') + 0.5 \mathcal{N}(x; \mu_2, \Sigma')$.
A stable split must ensure that the global statistical properties of the radiance field do not change abruptly at the moment of splitting. 
Mathematically, this is equivalent to matching the moments of the original distribution $p_{orig}$ and the new mixture $p_{mix}$.

\subsection{First Moment Matching}
The mean of the mixture is $\mathbb{E}[p_{mix}] = 0.5\mu_1 + 0.5\mu_2.$
To preserve the original position ($\mathbb{E}[p_{mix}] = \mu$), we must impose symmetry around $\mu$:
\begin{equation}
    \mu_1 = \mu + \Delta, \quad \mu_2 = \mu - \Delta.
\end{equation}
We choose the splitting direction $\Delta$ to be aligned with the principal eigenvector $\mathbf{v}_k$ of $\Sigma$, as this is the direction of maximum variance (uncertainty). Thus, $\Delta = \delta \mathbf{v}_k$.

\subsection{Second Moment Matching}
We require the covariance of the mixture to equal the original covariance $\Sigma$.
Using the law of total variance, the covariance of a mixture model is the sum of the average intra-component covariance and the inter-component covariance:
\begin{equation}
    \text{Cov}[p_{mix}] = \underbrace{\Sigma'}_{\text{Intra-cluster}} + \underbrace{\text{Var}[\{\mu_1, \mu_2\}]}_{\text{Inter-cluster}}.
\end{equation}

Substituting our symmetric means $\mu \pm \delta \mathbf{v}_k$:
\begin{equation}
\begin{split}
    \text{Var}[\{\mu_1, \mu_2\}] &= 0.5(\Delta)(\Delta)^T + 0.5(-\Delta)(-\Delta)^T = \Delta \Delta^T = \delta^2 \mathbf{v}_k \mathbf{v}_k^T.
\end{split}
\end{equation}

Let the original covariance be $\Sigma = \sum s_i^2 \mathbf{v}_i \mathbf{v}_i^T$ and the new covariance be isotropic relative to the original frame, $\Sigma' = \sum (s'_i)^2 \mathbf{v}_i \mathbf{v}_i^T$.
Equating $\text{Cov}[p_{mix}] = \Sigma$ along the principal axis $\mathbf{v}_k$:
\begin{equation}
    (s'_k)^2 + \delta^2 = s_k^2 \implies (s'_k)^2 = s_k^2 - \delta^2.
\end{equation}

However, in practice we observe that strict theoretical moment preservation can lead to optimization ambiguity due to the significant spatial overlap of the newly split kernels. 
To mitigate this, we introduce a contractive regularization to the scales. 
This explicitly reduces the mixture's total variance compared to the original distribution, ensuring the new primitives are sufficiently separated.

For the non-principal axes $j \neq k$, there is no mean displacement ($\Delta \cdot \mathbf{v}_j = 0$). 
We scale $s'_j = s_j \sqrt{1 - \eta^2}$ to maintain isotropic reduction relative to the split.
For the principal axis $k$, we apply the scale update $s'_k = s_k \sqrt{1 - \eta^2}$. This contractive bounding deliberately tightens the overall distribution to prevent popping artifacts while providing sufficient gradient separation for independent optimization.

\subsection{Optimal Choice of $\eta$}
We aim for the two new sub-Gaussians to cover the original extent while being distinct enough to optimize independently.
We consider the energy conservation of the probability density at the center ($x=0$).
The original PDF at the center is proportional to $P(0) \propto \frac{1}{s_k}$.
The new mixture PDF at the center is $P_{mix}(0) = \frac{1}{s'_k} \exp\left(-\frac{\delta^2}{2(s'_k)^2}\right)$.
We require the center density to remain continuous to avoid popping artifacts or density holes:
\begin{equation}
    \frac{1}{s_k} \approx \frac{1}{s_k \sqrt{1-\eta^2}} \exp\left(-\frac{(\eta s_k)^2}{2 s_k^2 (1-\eta^2)}\right).
\end{equation}
Simplifying this equation yields the transcendental relation:
\begin{equation}
    \sqrt{1-\eta^2} = \exp\left( -\frac{\eta^2}{2(1-\eta^2)} \right).
\end{equation}
The solution lies near $\eta \approx 0.45$. 
This empirical optimum balances center density preservation and providing sufficient gradient separation for the optimizer.

 \section{Prune Strategy of Geometric Deficiency Score}

In practice, primitives are first filtered by basic geometric and opacity constraints.
Among the remaining candidates, we define an inverse importance weight:
\begin{equation}
    w_i = \frac{1}{\epsilon + (1 - S_p(g_i))},
\end{equation}
which biases the pruning process toward primitives with lower $S_p$.
A fixed removal budget is then enforced by sampling primitives according to $w_i$ without replacement, and only sampled primitives that satisfy the pruning conditions are removed.

This probabilistic, budgeted pruning strategy avoids overly aggressive deterministic thresholding,
while preferentially eliminating hard-to-optimize, low-impact primitives.
As a result, the model maintains rendering fidelity while effectively reducing redundancy.

\section{Validation of the Gelbrich Distance Approximation}
\label{sec:gelbrich_validation}

In the implementation of our proposed OT aggregation algorithm, we utilize the Gelbrich distance, a lower bound on the exact Bures-Wasserstein distance, as a proxy for the transport cost. 
A critical consideration is whether this approximation remains tight under practical reconstruction conditions.

Mathematically, the Gelbrich distance provides a completely tight bound to the exact Bures-Wasserstein distance under the condition that the covariance matrices of the two target distributions commute. Specifically, for two given covariance matrices $\Sigma_i$ and $\Sigma_j$, this condition is satisfied when:
\begin{equation}
    \Sigma_i \Sigma_j = \Sigma_j \Sigma_i
    \label{eq:commutative_condition}
\end{equation}

In the context of our algorithm, this mathematical condition is inherently satisfied for the candidate pairs that are most critical to the aggregation process. Because our method performs the OT aggregation within localized, importance-balanced KD-Tree blocks, the candidate Gaussians evaluated for merging are spatially proximate and typically represent the exact same local surface patch.

Consequently, redundant Gaussians that fit the same local geometry naturally converge to have highly aligned principal axes. When the orientations of two such Gaussians are aligned, their covariance matrices become simultaneously diagonalizable. This simultaneous diagonalizability implies that the covariance matrices effectively commute, thus satisfying the condition defined in Equation~\ref{eq:commutative_condition}.

Therefore, we conclude that for the high-density redundant clusters where the optimal transport aggregation predominantly takes place, the Gelbrich distance serves as an exceptionally tight and reliable approximation of the exact Bures-Wasserstein distance.

\section{Detailed Experiment Results}
\label{der}

\paragraph{Detailed Implementation Setting}
We apply a unified preprocessing pipeline to the four datasets used in our study, namely Waymo~\cite{Sun_2020_CVPRwaymo}, Mip-NeRF 360~\cite{barron2022mip}, Tanks $\&$ Temples~\cite{Knapitsch2017}, and Deep Blending~\cite{hedman2018deep}, to make them compatible with the input requirements of the 3DGS~\cite{kerbl20233d} framework.
The data process strictly follows the 3D Gaussian Splatting.
Specifically, we use COLMAP to perform structure-from-motion (SFM) reconstruction on the raw image sequences. 
This process yields sparse point clouds, camera poses, and point correspondences across images.

For the Waymo Dataset, we focus on images captured by the driving vehicles through three cameras. 
Given the challenges of wide-baseline viewpoints in autonomous driving scenarios, we carefully selected six sequences that showed great challenges like turning around and sharp changing during reconstruction. 
For the Mip-NeRF 360 dataset, we use seven representative scenes, which cover a range of indoor and outdoor environments with varying geometry and material complexity, allowing us to evaluate generalization performance across diverse conditions.
For the Tanks $\&$ Temples and Deep Blending dataset, we follow the standard protocols used in prior work, adopting the same training/testing splits and evaluation metrics to ensure comparability.

\subsection{Detailed Comparison Results}
\label{detailed results}

\begin{table*}[ht]
\vspace{-10pt}
\small
\caption{\textbf{Quantitative average GPU peak memory (GB) results to previous methods.}}
\label{gpu}
\center
\sc
\setlength{\tabcolsep}{6pt} 
\begin{tabular}{l|cccc}
\toprule
Dataset & {Mip-NeRF 360}     &{Deep Blending}    &{Tanks $\&$ Temples}             &{Waymo}              \\
\midrule
3DGS            & 9.41 & 8.26 & 4.79 & 21.28 \\
DashGaussian    & 9.01 & 7.74 & 4.48 & 21.19 \\
SPEEDY-SPLAT    & 6.71 & 5.25 & 2.77 & 14.56 \\
GHAP            & 9.05 & 8.11 & 4.74 & 19.72 \\
mini-splatting  & 7.22 & 6.27 & 4.65 & 20.53 \\
taming-3dgs     & 5.23 & 3.93 & 2.49 & 14.59 \\
FastGS          & 4.71 & 3.88 & 2.42 & 15.68 \\
\midrule
MMGS (Ours)     & 5.26 & 4.08 & 2.75 & 14.46 \\
\bottomrule
\end{tabular}
\end{table*}

In this section, we present comprehensive quantitative and qualitative comparisons between our proposed approach (MMGS and its variant MMGS-B) and several recent state-of-the-art 3D Gaussian Splatting (3DGS) baselines. The evaluations are conducted across multiple widely used benchmarks, including the Mip-NeRF 360~\cite{barron2022mip}, Waymo~\cite{Sun_2020_CVPRwaymo}, Tanks \& Temples~\cite{Knapitsch2017}, and Deep Blending~\cite{hedman2018deep} datasets.

\begin{table}[htb]
\small
\caption{\textbf{Rendering speed (FPS) comparison.} Best results are \textbf{bolded}.}
\label{tab:fps_only}
\center
\sc
\begin{tabular}{l|ccc}
\toprule
Method & Mip-NeRF 360 & Tanks $\&$ Temples & Deep Blending \\
\midrule
3DGS           & 86  & 115 & 87  \\
LightGaussian  & 165 & 217 & 174 \\
Compact-3DGS   & 152 & 197 & 193 \\
RadSplat       & 176 & 225 & 210 \\
DashGaussian   & 106 & 148 & 106 \\
Speedy-splat   & 191 & 186 & 235 \\
GHAP           & 316 & 354 & 333 \\
Mini-splatting & 308 & 372 & 344 \\
Taming-3dgs    & 160 & 231 & 226 \\
FastGS         & 327 & 389 & 359 \\
\midrule
MMGS (ours)    & \textbf{355} & \textbf{397} & \textbf{368} \\
MMGS-B (ours)  & 293 & 339 & 301 \\
\bottomrule
\end{tabular}
\vspace{-10pt}
\end{table}

\paragraph{Memory Consumption and Rendering Efficiency.} 
One of the core advantages of our method is its exceptional hardware efficiency. As summarized in Table~\ref{gpu}, MMGS significantly reduces the average peak GPU memory requirements across all evaluated datasets compared to the vanilla 3DGS and other concurrent works such as DashGaussian and GHAP. Furthermore, Table~\ref{tnt1} details the training time, Gaussian primitive count ($N_{GS}$), and rendering speed (FPS). Notably, MMGS achieves real-time rendering speeds exceeding 350 FPS on multiple scenes while requiring substantially lower training times, demonstrating its highly optimized pipeline.

\paragraph{Model Compression and Primitive Reduction.} 
To further illustrate our model's compactness, we provide a scene-by-scene breakdown of the total number of Gaussians. As shown in the lower part of Table~\ref{3603} for the Mip-NeRF 360 dataset and Table~\ref{waymo2} for the large-scale Waymo dataset, MMGS dramatically prunes redundant Gaussians. For instance, on the \textit{bicycle} scene, MMGS reduces the primitive count from 4.90M to merely 0.51M, while similar magnitude reductions are consistently observed in challenging unbounded outdoor driving scenarios.

\paragraph{Quantitative Rendering Quality.} 
Despite the drastic reduction in both memory footprint and the number of primitives, our method does not compromise on novel view synthesis quality. Detailed per-scene PSNR metrics for the Mip-NeRF 360 and Waymo datasets are provided in the upper part of Table~\ref{3602} and Table~\ref{waymo1}, respectively. Additionally, comprehensive metrics including PSNR, SSIM, and LPIPS on the Tanks \& Temples and Deep Blending datasets are reported in Table~\ref{tnt2}. The results indicate that MMGS-B consistently achieves highly competitive, and in many cases superior, rendering fidelity compared to models with much heavier memory footprints.

\begin{figure*}[ht]
  \begin{center}
    \centerline{\includegraphics[width=1.0\columnwidth, trim=5.6cm 3.6cm 0cm 0cm, clip]{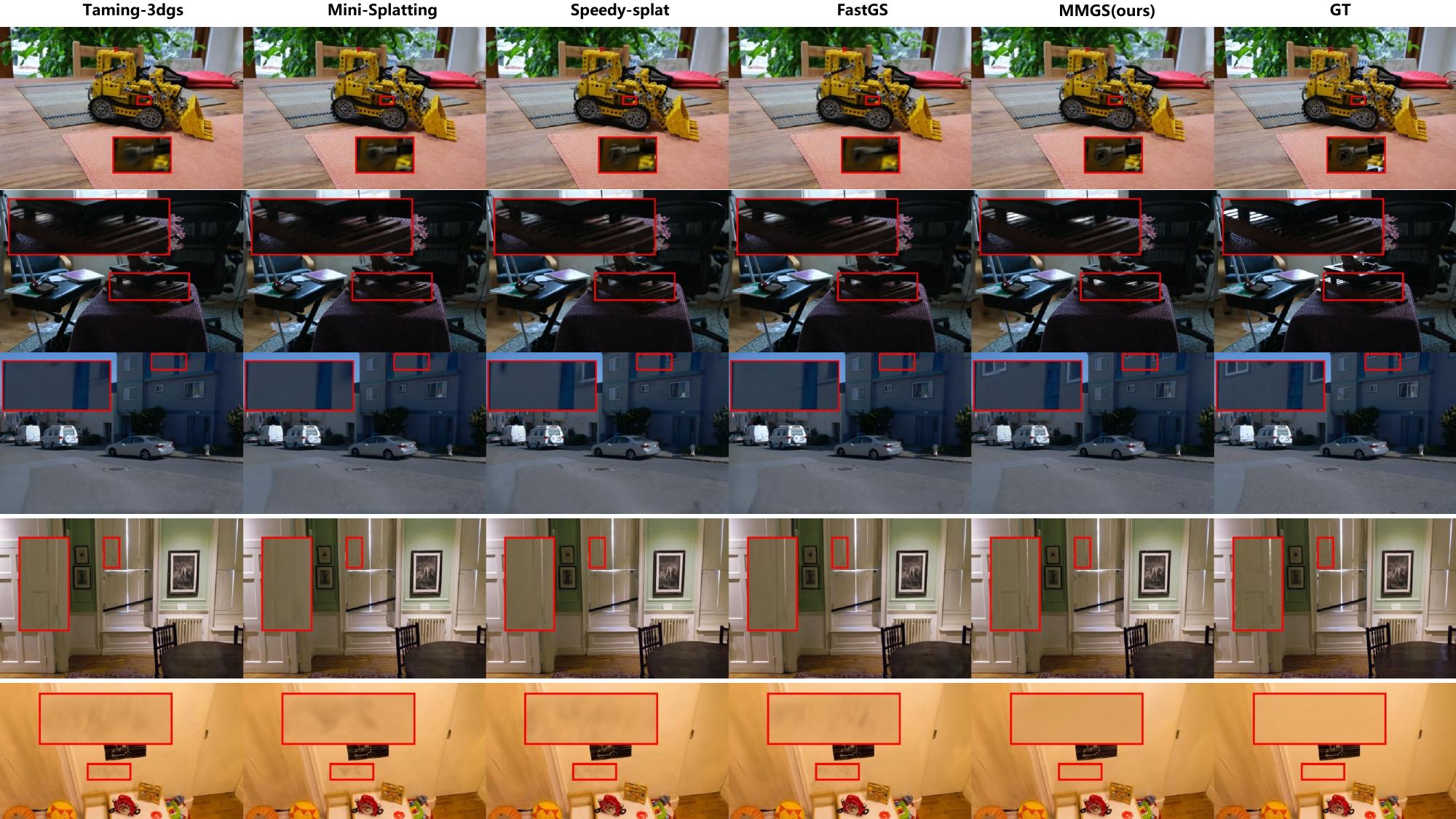}}
    \caption{\textbf{Comparison Results.} Visual differences are highlighted with red insets for better clarity. Our approach consistently outperforms other models on different scenes, demonstrating advantages in challenging scenarios. Best viewed in color.}
    \label{e3}
  \end{center}
  \vspace{-20pt}
\end{figure*}

\begin{figure*}[ht]
  \begin{center}
    \centerline{\includegraphics[width=1.0\columnwidth,trim=5.6cm 7.9cm 0cm 0cm, clip]{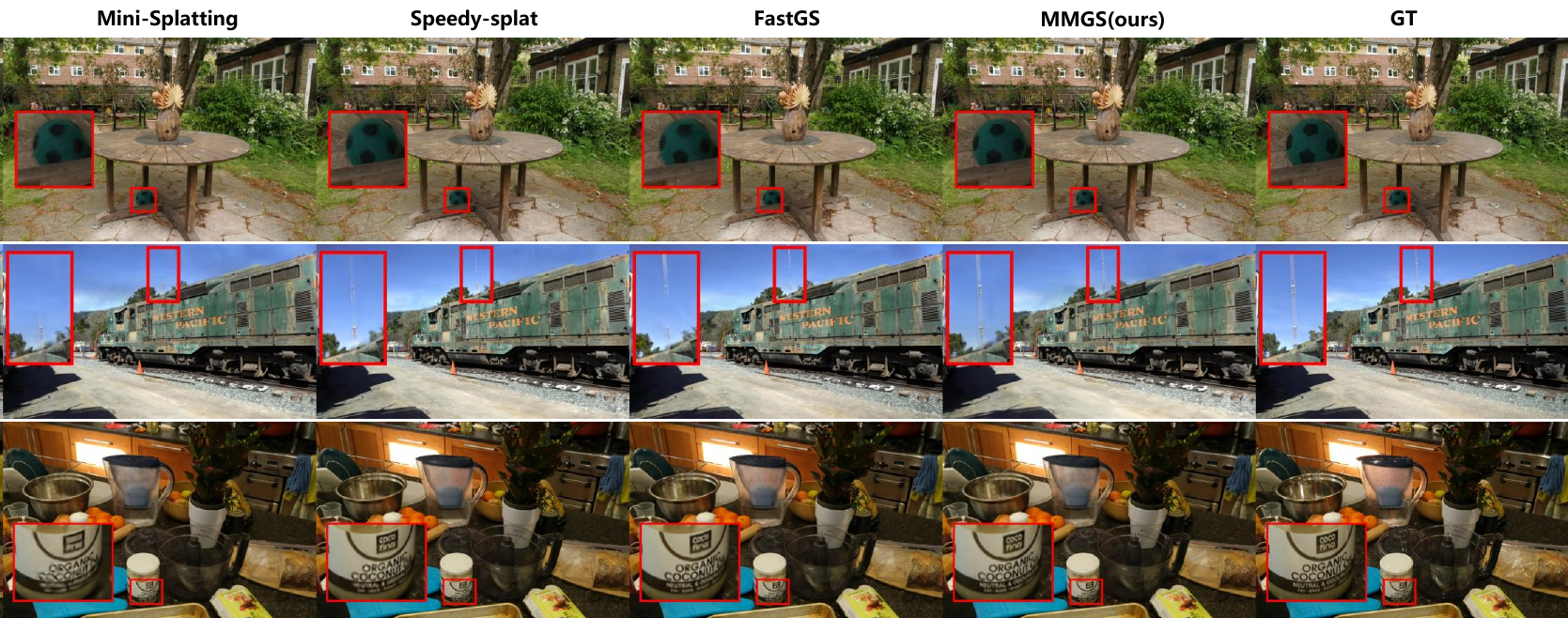}}
    \caption{\textbf{Comparison Results.} Visual differences are highlighted with red insets for better clarity. Our approach consistently outperforms other models on different scenes, demonstrating advantages in challenging scenarios. Best viewed in color.}
    \label{e4}
  \end{center}
\end{figure*}

\paragraph{Qualitative Evaluation.} 
Finally, visual comparisons against baseline methods are illustrated in Figure~\ref{e3} and Figure~\ref{e4}. As highlighted by the red insets, our approach is capable of preserving fine high-frequency details and mitigating prominent artifacts that commonly appear in baseline renderings. MMGS consistently outperforms other models across different scenes, demonstrating distinct advantages in reconstructing challenging structures and complex backgrounds.

\begin{table*}[htb]
\small
\caption{\textbf{Quantitative PSNR results to previous methods on Mip-NeRF 360~\cite{barron2022mip} dataset.} }
\label{3602}
\center
\sc
\setlength{\tabcolsep}{6pt} 
\begin{tabular}{l|ccccccc}
\toprule
\multicolumn{1}{l|}{Scene}                                     & \multicolumn{1}{c}{bicycle} & \multicolumn{1}{c}{bonsai} & \multicolumn{1}{c}{counter} & \multicolumn{1}{c}{garden} & \multicolumn{1}{c}{kitchen} & \multicolumn{1}{c}{room} & \multicolumn{1}{c}{stump} \\
\midrule
3DGS                                 & {\color[HTML]{1F1F1F} 25.22}         & {\color[HTML]{1F1F1F} 32.23}        & {\color[HTML]{1F1F1F} 29.06}         & {\color[HTML]{1F1F1F} 27.37}        & {\color[HTML]{1F1F1F} 31.36}         & {\color[HTML]{1F1F1F} 31.59}      & {\color[HTML]{1F1F1F} 26.61}       \\
DashGaussian                                    & 25.47                                & 31.76                               & 28.95                                & 27.56                               & 31.45                                & 31.60                             & 27.08                              \\
Speedy-splat & 25.57                                & 31.59                               & 28.69                                & 27.50                                & 30.56                                & 31.97                             & 26.36                              \\
GHAP                                            & 25.37                                & 30.97                               & 28.38                                & 27.15                               & 30.44                                & 30.81                             & 26.53                              \\
mini-splatting                                 & 25.66                                & 31.41                               & 28.59                                & 27.24                               & 31.34                                & 31.41                             & 27.50                              \\
taming-3dgs                                     & 24.83                                & 31.85                               & 28.56                                & 27.28                               & 31.04                                & 31.30                             & 26.02                              \\
FastGS                                          & 25.47                                & 31.50                               & 28.59                                & 27.12                               & 31.18                                & 31.67                             & 26.87                              \\
\midrule
MMGS (Ours)                                  & 25.48                                & 31.33                               & 28.62                                & 27.14                               & 31.17                                & 31.68                             & 26.87                              \\
MMGS-B   (Ours)                               & 25.72                                & 31.54                               & 28.92                                & 27.47                               & 31.40                                & 32.03                             & 27.14                  \\
\bottomrule
\end{tabular}
\end{table*}

\begin{table*}[htb]
\small
\caption{\textbf{Quantitative Gaussian number to previous methods on Mip-NeRF 360~\cite{barron2022mip} dataset.}}
\label{3603}
\center
\sc
\setlength{\tabcolsep}{6pt} 
\begin{tabular}{l|ccccccc}
\toprule
\multicolumn{1}{l|}{Scene}                                     & \multicolumn{1}{c}{bicycle} & \multicolumn{1}{c}{bonsai} & \multicolumn{1}{c}{counter} & \multicolumn{1}{c}{garden} & \multicolumn{1}{c}{kitchen} & \multicolumn{1}{c}{room} & \multicolumn{1}{c}{stump} \\
\midrule
3DGS                                & {\color[HTML]{1F1F1F} 4.90M}         & {\color[HTML]{1F1F1F} 1.06M}        & {\color[HTML]{1F1F1F} 1.07M}         & {\color[HTML]{1F1F1F} 4.08M}        & {\color[HTML]{1F1F1F} 1.58M}         &  1.29M      & 4.32M       \\
DashGaussian                                    & 4.68M                                & 0.86M                               & 0.84M                                & 3.22M                               & 1.22M                                & 1.17M                             & 3.51M                              \\
Speedy-splat & 1.15M                                & 0.25M                               & 0.19M                                & 0.91M                               & 0.22M                                & 0.21M                             & 0.74M                              \\
GHAP                                            & 0.94M                                & 0.21M                               & 0.21M                                & 0.73M                               & 0.31M                                & 0.25M                             & 0.82M                              \\
mini-splatting                                  & 0.53M                                & 0.35M                               & 0.41M                                & 0.59M                               & 0.43M                                & 0.39M                             & 0.60M                              \\
taming-3dgs                                     & 0.81M                                & 0.41M                               & 0.31M                                & 1.90M                               & 0.48M                                & 0.23M                             & 0.48M                              \\
FastGS                                          & 0.85M                                & 0.25M                               & 0.20M                                & 0.51M                               & 0.28M                                & 0.21M                             & 0.62M                              \\
\midrule
MMGS (Ours)                                   & 0.51M                                & 0.19M                               & 0.16M                                & 0.45M                               & 0.22M                                & 0.16M                             & 0.34M                              \\
MMGS-B (Ours)                               & 0.99M                                & 0.37M                               & 0.32M                                & 0.91M                               & 0.46M                                & 0.33M                             & 0.61M     \\
\bottomrule
\end{tabular}
\end{table*}

\begin{table*}[htb]
\small
\caption{\textbf{Quantitative PSNR comparison results to previous methods on Waymo~\cite{Sun_2020_CVPRwaymo} dataset.}}
\label{waymo1}
\center
\sc
\setlength{\tabcolsep}{6pt} 
\begin{tabular}{l|ccccccc}
\toprule
\multicolumn{1}{l|}{Scene}    & \multicolumn{1}{c}{100613} & \multicolumn{1}{c}{117240} & \multicolumn{1}{c}{130854} & \multicolumn{1}{c}{148697} & \multicolumn{1}{c}{152706} & \multicolumn{1}{c}{158686} \\
\midrule
3DGS                                 & {\color[HTML]{1F1F1F} 31.64}           & {\color[HTML]{1F1F1F} 31.40}        & {\color[HTML]{1F1F1F} 24.62}        & {\color[HTML]{1F1F1F} 24.64}        & {\color[HTML]{1F1F1F} 21.40}        & {\color[HTML]{1F1F1F} 27.69}        \\
DashGaussian                                    & 28.29                               & 30.59                               & 24.36                               & 23.20                               & 20.10                               & 27.25                               \\
Speedy-splat & 28.02                               & 30.35                               & 23.70                                & 22.58                               & 19.48                               & 26.72                               \\
GHAP                                            & 30.60                               & 30.52                               & 24.39                               & 24.29                               & 21.52                               & 27.21                               \\
mini-splatting                                  & 25.95                               & 26.32                               & 21.18                               & 20.07                               & 20.54                               & 26.44                               \\
taming-3dgs                                     & 31.07                               & 30.38                               & 24.75                               & 24.58                               & 22.50                               & 27.39                               \\
fastgs       & 31.39                               & 31.36       & 24.32       & 24.82       & 21.91       & 27.48       \\
\midrule
MMGS (Ours)                                   & 31.21                               & 31.27                               & 24.54                               & 24.99                               & 22.11                               & 27.49                               \\
MMGS-B (Ours)                               & 31.54                               & 31.54                               & 24.89                               & 25.30                               & 22.40                               & 27.80        \\
\bottomrule
\end{tabular}
\end{table*}

\begin{table*}[htb]
\vspace{-45pt}
\small
\caption{\textbf{Quantitative Gaussian number results to previous methods on Waymo~\cite{Sun_2020_CVPRwaymo} dataset.}}
\label{waymo2}
\center
\sc
\setlength{\tabcolsep}{6pt} 
\begin{tabular}{l|ccccccc}
\toprule
\multicolumn{1}{l|}{Scene}    & \multicolumn{1}{c}{100613} & \multicolumn{1}{c}{117240} & \multicolumn{1}{c}{130854} & \multicolumn{1}{c}{148697} & \multicolumn{1}{c}{152706} & \multicolumn{1}{c}{158686} \\
\midrule
3DGS                                   & {\color[HTML]{1F1F1F} 2.52M}        & {\color[HTML]{1F1F1F} 1.71M}        & {\color[HTML]{1F1F1F} 1.36M}        & {\color[HTML]{1F1F1F} 2.02M}        & {\color[HTML]{1F1F1F} 0.20M}         & {\color[HTML]{1F1F1F} 3.27M}        \\
DashGaussian                                    & 1.34M                               & 1.36M                               & 1.34M                               & 1.21M                               & 0.16M                               & 2.31M                               \\
Speedy-splat & 0.52M                               & 0.46M                               & 0.38M                               & 0.47M                               & 0.03M                               & 1.03M                               \\
GHAP                                            & 0.50M                               & 0.36M                               & 0.31M                               & 0.44M                               & 0.04M                               & 0.66M                               \\
mini-splatting                                  & 0.31M                               & 0.24M                               & 0.34M                               & 0.21M                               & 0.27M                               & 0.41M                               \\
taming-3dgs                                     & 0.48M                               & 0.29M                               & 0.31M                               & 0.37M                               & 0.13M                               & 0.59M                               \\
\rowcolor[HTML]{FFFFFF} 
FastGS        & 0.43M                               & 0.32M                               & 0.43M                               & 0.59M                               & 0.09M                               & 0.65M                               \\
\midrule
MMGS (OURS)                                            & 0.22M                               & 0.20M                               & 0.23M                               & 0.30M                               & 0.06M                               & 0.31M                               \\
MMGS-B (Ours)                               & 0.52M                               & 0.45M                               & 0.53M                               & 0.69M                               & 0.14M                               & 0.81M               \\
\bottomrule
\end{tabular}
\vspace{-10pt}
\end{table*}

\begin{table*}[htb]
\small
\caption{\textbf{Quantitative results on Tanks $\&$ Temples~\cite{Knapitsch2017} and Deep Blending~\cite{hedman2018deep} dataset.} }
\label{tnt1}
\center
\sc
\setlength{\tabcolsep}{2pt} 
\resizebox{1.0\linewidth}{!}{
\begin{tabular}{l|cccccccccccc}
\toprule
Dataset                                 & \multicolumn{6}{c|}{Deep Blending}      & \multicolumn{6}{c}{Tanks $\&$ Temples}                                                                                                                                                                                            \\
Scene                                   & \multicolumn{3}{c|}{{\color[HTML]{1F1F1F} drjohnson}}                                                      & \multicolumn{3}{c|}{{\color[HTML]{1F1F1F} playroom}}                                                       & \multicolumn{3}{c|}{train}                                                                                 & \multicolumn{3}{c}{truck}                                                                                 \\
Metric                                  & \multicolumn{1}{c}{Time} & \multicolumn{1}{c}{\textbf{$N_{GS}$}} & \multicolumn{1}{c|}{FPS} & \multicolumn{1}{c}{Time} & \multicolumn{1}{c}{\textbf{$N_{GS}$}} & \multicolumn{1}{c|}{FPS} & \multicolumn{1}{c}{Time} & \multicolumn{1}{c}{\textbf{$N_{GS}$}} & \multicolumn{1}{c|}{FPS} & \multicolumn{1}{c}{Time} & \multicolumn{1}{c}{\textbf{$N_{GS}$}} & \multicolumn{1}{c}{FPS} \\
\midrule
3DGS & \cellcolor[HTML]{FFFFFF}34.13 & \cellcolor[HTML]{FFFFFF}3.08M          & \cellcolor[HTML]{FFFFFF}73       & \cellcolor[HTML]{FFFFFF}26.23          & \cellcolor[HTML]{FFFFFF}1.85M          & \cellcolor[HTML]{FFFFFF}102      & 16.05                                  & 1.09M                                  & 107                              & 19.55                                  & 2.05M                                  & 124                              \\
DashGaussian                            & 8.04                                   & 2.35M                                  & 109                              & 6.33                                   & 1.53M                                  & 102                              & 7.34                                   & 0.98M                                  & 137                              & 7.11                                   & 1.43M                                  & 159                              \\
\rowcolor[HTML]{FFFFFF} 
Speedy-splat                            & 22.24                                  & 0.61M                                  & 231                           & 18.15                                  & 0.36M                                  & 237                           & 12.47                                  & 0.17M                                  & 129                           & 14.54                                  & 0.26M                                  & 242                          \\
GHAP                                    & 22.91                                  & 0.61M                                  & 322                              & 17.62                                  & 0.37M                                  & 342                              & 11.12                                  & 0.22M                                  & 347                              & 13.15                                  & 0.41M                                  & 360                              \\
mini-splatting                          & 23.26                                  & 0.60M                                  & 336                              & 20.42                                  & 0.51M                                  & 352                              & 16.43                                  & 0.28M                                  & 409                              & 16.83                                  & 0.33M                                  & 334                              \\
taming-3dgs                             & 4.45                                   & 0.40M                                  & 217                              & 3.66                                   & 0.18M                                  & 235                              & 5.08                                   & 0.37M                                  & 207                              & 4.61                                   & 0.27M                                  & 255                              \\
FastGS                                  & 3.16                                   & 0.39M                                  & 357                              & 2.79                                   & 0.25M                                  & 385                              & 2.79                                   & 0.22M                                  & 377                              & 2.86                                   & 0.28M                                  & 400                              \\
\midrule
MMGS (Ours)                           & 3.19                                   & 0.28M                                  & 367                              & 2.91                                   & 0.20M                                  & 368                              & 2.77                                   & 0.18M                                  & 397                              & 2.87                                   & 0.22M                                  & 395                              \\
MMGS-B (Ours)                       & 4.82                                   & 0.52M                                  & 307                              & 4.36                                   & 0.40M                                  & 295                              & 3.93                                   & 0.32M                                  & 351                              & 4.12                                   & 0.35M                                  & 327               \\
\bottomrule
\end{tabular}}
\vspace{-5pt}
\end{table*}

\begin{table*}[ht]
\small
\caption{\textbf{Quantitative results on Tanks $\&$ Temples~\cite{Knapitsch2017} and Deep Blending~\cite{hedman2018deep} dataset.} }
\label{tnt2}
\center
\sc
\setlength{\tabcolsep}{2pt} 
\resizebox{1.0\linewidth}{!}{
\begin{tabular}{l|ccc|ccc|ccc|ccc}
\toprule
Dataset                                 & \multicolumn{6}{c|}{Deep Blending}      & \multicolumn{6}{c}{Tanks $\&$ Temples}                                                                                                                                                                                            \\
Scene                                   & \multicolumn{3}{c|}{{\color[HTML]{1F1F1F} drjohnson}}                                                      & \multicolumn{3}{c|}{{\color[HTML]{1F1F1F} playroom}}                                                       & \multicolumn{3}{c|}{train}                                                                                 & \multicolumn{3}{c}{truck}                                                                                 \\
Metric                                  & \multicolumn{1}{c}{PSNR}                             & \multicolumn{1}{c}{SSIM}                    & \multicolumn{1}{c|}{LPIPS}                   & \multicolumn{1}{c}{PSNR}                    & \multicolumn{1}{c}{SSIM}                    & \multicolumn{1}{c|}{LPIPS}                   & \multicolumn{1}{c}{PSNR} & \multicolumn{1}{c}{SSIM} & \multicolumn{1}{c|}{LPIPS} & \multicolumn{1}{c}{PSNR} & \multicolumn{1}{c}{SSIM} & \multicolumn{1}{c}{LPIPS} \\
\midrule
3DGS & 29.39 &  0.903 & \cellcolor[HTML]{FFFFFF}{\color[HTML]{1F1F1F} 0.248} & \cellcolor[HTML]{FFFFFF}{\color[HTML]{1F1F1F} 30.08} & \cellcolor[HTML]{FFFFFF}{\color[HTML]{1F1F1F} 0.903} & \cellcolor[HTML]{FFFFFF}{\color[HTML]{1F1F1F} 0.252} & {\color[HTML]{1F1F1F} 22.16}      & {\color[HTML]{1F1F1F} 0.818}      & {\color[HTML]{1F1F1F} 0.208}       & {\color[HTML]{1F1F1F} 25.35}      & {\color[HTML]{1F1F1F} 0.881}      & {\color[HTML]{1F1F1F} 0.153}       \\
DashGaussian                            & 29.17                                                         & 0.900                                                & 0.252                                                & 30.04                                                & 0.904                                                & 0.246                                                & 22.10                             & 0.813                             & 0.210                              & 25.79                             & 0.882                             & 0.151                              \\
\rowcolor[HTML]{FFFFFF} 
Speedy-splat                            & 29.15                                                         & 0.902                                               & 0.254                                               & 30.12                                                & 0.908                                               & 0.257                                               & 21.73                             & 0.772                            & 0.289                             & 25.18                             & 0.867                            & 0.189                             \\
GHAP                                    & 29.37                                                         & 0.902                                                & 0.253                                                & 30.10                                                & 0.903                                                & 0.257                                                & 21.43                             & 0.791                             & 0.253                              & 24.95                             & 0.865                             & 0.181                              \\
mini-splatting                          & 29.57                                                         & 0.905                                                & 0.246                                                & 30.42                                                & 0.908                                                & 0.240                                                & 21.65                             & 0.809                             & 0.223                              & 25.21                             & 0.878                             & 0.139                              \\
taming-3dgs                             & 29.47                                                         & 0.900                                                & 0.269                                                & 30.20                                                & 0.898                                                & 0.278                                                & 22.49                             & 0.803                             & 0.238                              & 25.12                             & 0.863                             & 0.186                              \\
FastGS                                  & 29.45                                                         & 0.900                                                & 0.260                                                & 30.63                                                & 0.907                                                & 0.254                                                & 22.43                             & 0.799                             & 0.246                              & 25.69                             & 0.871                             & 0.178                              \\
\midrule
MMGS (Ours)                           & 29.68                                                         & 0.901                                                & 0.262                                                & 30.68                                                & 0.906                                                & 0.258                                                & 22.48                             & 0.803                             & 0.244                              & 25.66                             & 0.872                             & 0.172                              \\
MMGS-B   (Ours)                       & 29.79                                                         & 0.908                                                & 0.249                                                & 30.74                                                & 0.908                                                & 0.247                                                & 22.62                             & 0.817                             & 0.212                              & 25.82                             & 0.884                             & 0.148                             \\
\bottomrule
\end{tabular}}
\vspace{-5pt}
\end{table*}

\section{Ethics Statement}
This work adheres to the NIPS Code of Ethics. In this study, no human subjects or animal experimentation was involved. All datasets used were sourced in compliance with relevant usage guidelines, ensuring no violation of privacy. We have taken care to avoid any biases or discriminatory outcomes in our research process. No personally identifiable information was used, and no experiments were conducted that could raise privacy or security concerns. We are committed to maintaining transparency and integrity throughout the research process.

\section{Reproducibility Statement}
We have made every effort to ensure that the results presented in this paper are reproducible. All code and datasets will be made publicly available after the paper is accepted to facilitate replication and verification. 
We believe these measures will enable other researchers to reproduce our work and further advance the field.

\clearpage
\newpage

\end{document}